\RequirePackage[svgnames]{xcolor}
\documentclass[11pt,letterpaper]{mystyle}
\usepackage[comma,authoryear,compress]{natbib}
\usepackage[T1]{fontenc}
\usepackage[utf8]{inputenc}
\usepackage{amsmath,amssymb,amsthm,bm}
\usepackage{graphicx}
\usepackage{booktabs,multirow,array,makecell}
\usepackage{xcolor,colortbl}
\usepackage{adjustbox}
\usepackage{float}
\usepackage{placeins}
\usepackage{pifont}
\usepackage{hyperref}
\usepackage{url}
\usepackage{xurl}
\usepackage{eso-pic}
\usepackage[nameinlink,noabbrev]{cleveref}

\definecolor{PaperInk}{HTML}{293F59}
\definecolor{PaperPanel}{HTML}{F3F6FA}
\definecolor{TableGroup}{HTML}{EAF0F6}
\definecolor{TableSubtle}{HTML}{F7FAFC}
\definecolor{TableRule}{HTML}{CBD5E1}
\colorlet{TinaCrimson}{PaperInk}
\renewcommand{\titlefont}{\raggedright\color{PaperInk}\normalfont\fontsize{19}{23}\selectfont}
\tcbset{titlebox/.append style={colback=PaperPanel,colframe=PaperPanel}}

\hypersetup{
  colorlinks=true,
  allcolors=YaleBlue,
  pdftitle={Recall or Recheck? Evaluating Memory-Augmented Agents in Evolving Mobile Environments}
}

\newtheorem{definition}{Definition}[section]

\newcommand{\cmark}{\ding{51}}
\newcommand{\xmark}{\ding{55}}

\newcommand{\tablestyle}{\small\renewcommand{\arraystretch}{1.14}\setlength{\tabcolsep}{5pt}}
\newcommand{\tablegroup}[2]{\multicolumn{#1}{@{}>{\columncolor{TableGroup}}l@{}}{\rule[-0.55ex]{0pt}{2.65ex}\hspace{0.55em}\textcolor{PaperInk}{\textbf{#2}}}\\}

\title{Recall or Recheck? Evaluating Memory-Augmented Agents in Evolving Mobile Environments}
\runningtitle{Recall or Recheck? Evaluating Memory-Augmented Agents in Evolving Mobile Environments}

\renewcommand\Affilfont{\centering\normalfont\fontsize{9}{13}\selectfont}
\makeatletter
\renewcommand\AB@affilsepx{\hspace{0.75em}\protect\Affilfont}
\makeatother
\author[ ]{Qiyi Wang$^{1,\ddagger}$, Weigao Sun$^{2,\dagger}$, Junpeng Zhao$^{3}$, Keda Li$^{1}$, Junhan Shi$^{4}$, Xiao Lin$^{5}$, Yiran Zhong$^{2,\dagger}$, Zifeng Cheng$^{1,\dagger}$, Zhiwei Jiang$^{1,\dagger}$, Qing Gu$^{1}$, Steven Hoi$^{2}$}
\affil[1]{Nanjing University}
\affil[2]{Tongyi Lab, Alibaba Group}
\affil[3]{Fudan University}
\affil[4]{Tsinghua University}
\affil[5]{Tongji University}
\fancypagestyle{firststyle}{\fancyhf{}\fancyfoot[L]{\footerfont\itshape $^{\dagger}$Corresponding authors.\quad $^{\ddagger}$Work done during internship at Tongyi Lab, Alibaba Group.}}
\date{}
\begin{document}
\AddToShipoutPictureFG*{
  \AtPageUpperLeft{\put(\LenToUnit{1.9cm},\LenToUnit{-1.55cm}){\includegraphics[height=0.9cm]{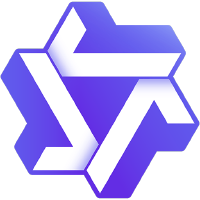}}\put(\LenToUnit{1.9cm},\LenToUnit{-1.8cm}){\rule{\textwidth}{0.5pt}}}}

\begin{abstract}
Long-term memory enables agents to retain past experience, yet existing evaluations largely treat accumulated memory as a self-contained source of truth.
In an evolving mobile environment, however, historical records may disappear from the current application state while recent updates remain absent from memory.
We introduce \textbf{$\bm{R^2}$-MobileMemory}, an executable benchmark of 513 tasks across ten year-long bilingual mobile worlds that separates persistent memory formed earlier from the current application state, testing when agents should recall, recheck, or combine both sources.
Across three agents and eight memory backends, persistent memory and device search are complementary, raising Qwen3.8-Max task success from 29.2\% without memory and 18.1\% without device search to 56.3\% when combined.
Strong agents query memory on almost every task that needs earlier records, and most backends store over 90\% of those records, yet agents recover only about half of them even on the strongest backends.
Task difficulty grows with the number of records a task requires rather than their age, and for strong agents most of this difficulty disappears once those records are provided.
Simple dense retrieval performs on par with more costly LLM-based backends.
Forming memory up to evaluation time helps tasks about the current state but hurts tasks that need earlier records, leaving overall success nearly unchanged.
Together, these results show that memory-augmented agents depend not only on what memory retains, but also on deciding when to recall and on retrieving complete information from memory and an evolving environment.
\end{abstract}

\maketitle
\vspace{3mm}

\section{Introduction}

As large language models (LLMs) evolve from one-shot assistants into persistent agents, their decisions increasingly depend on information accumulated across prior interactions.
Long-term memory provides this continuity by preserving user-specific facts, prior decisions, and relevant experience beyond the current context~\citep{zhong2024memorybank,packer2023memgpt,shinn2023reflexion,zhao2024expel,hu2025memory}.
Existing evaluations have expanded from testing whether agents can recall information from long interaction histories to whether retained experience can support reasoning, tool use, and multi-step behavior~\citep{maharana2024evaluating,wu2024longmemeval,tavakoli2026beyond,shen2026mem2actbench,he2026memoryarena}.
Yet these evaluations largely treat accumulated memory or history as the primary source of information for downstream decisions~\citep{chai2026smmbench,bei2026mem,he2026memoryarena,liu2026worldmemarena}.
Because memory is formed from past observations and the world in which an agent acts keeps changing, relevant historical information may disappear from the current state, and newer observations may supersede what memory retains.

\begin{figure}[t]
\centering
\includegraphics[width=\textwidth]{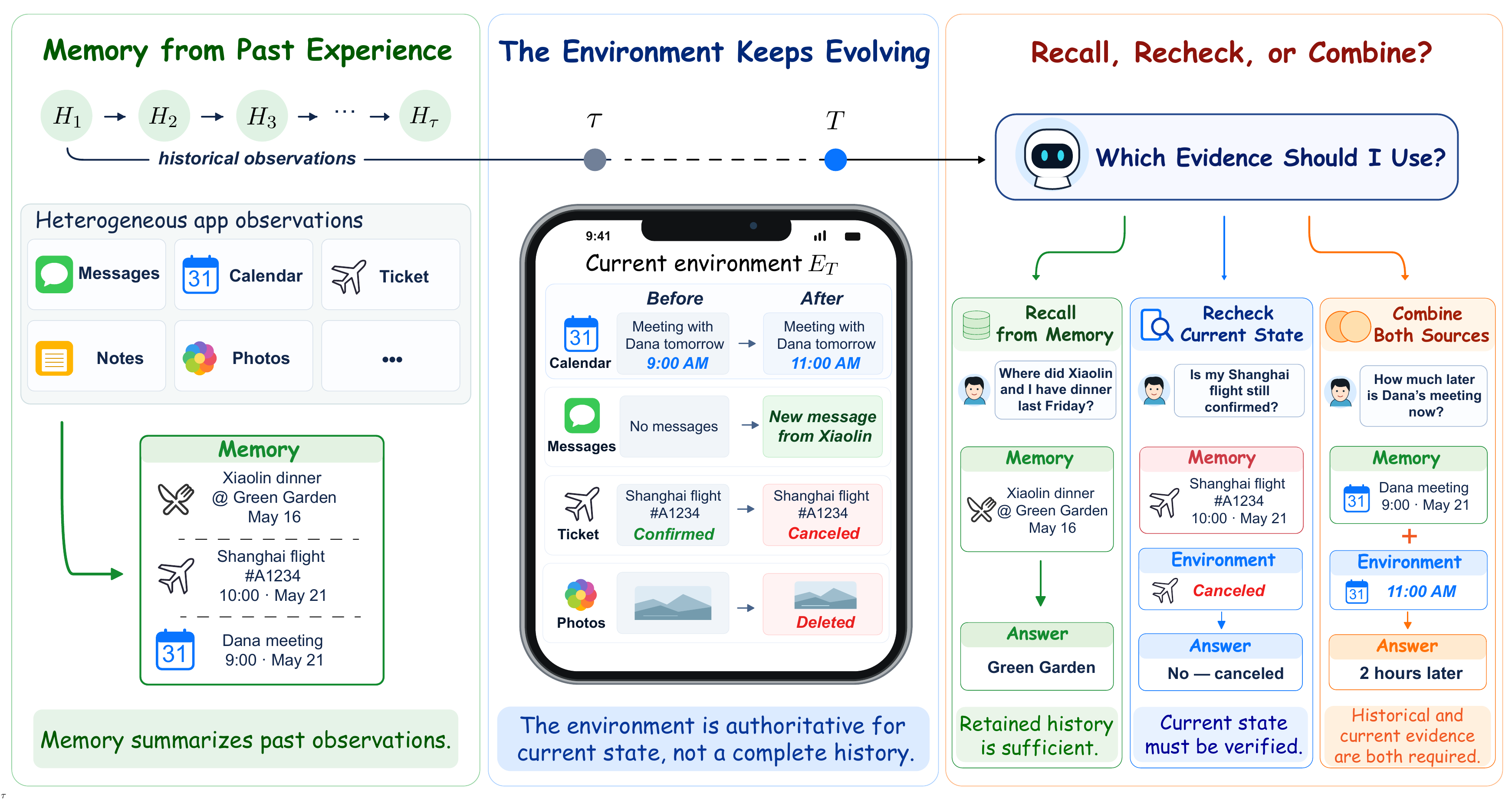}
\caption{\textbf{Recall or recheck in evolving mobile environments.}
Long-term memory is formed from historical observations, while the phone continues to evolve: new information may appear and old records may disappear.
At evaluation time, an agent must recall retained experience, recheck the current application state, or combine both to gather the information needed to complete the executable task.}
\label{fig:recall_or_recheck}
\end{figure}

Mobile environments make this challenge particularly clear. 
Smartphones are continuously evolving personal information spaces, where messages arrive, plans are revised, files are replaced, orders progress, and records may eventually disappear. 
At the same time, task-relevant information is fragmented across applications, modalities, and time. Recent work has advanced long-horizon personal memory and personalized mobile assistance~\citep{deng2026mobilemem,chen2026knowu,nie2026pspa}, while increasingly realistic mobile benchmarks support cross-session interaction, GUI execution, and tool-augmented workflows~\citep{rawles2025androidworld,kong2026mobileworld,zhu2026mobilepa,liu2026memgui}.
Yet these settings rarely treat persistent memory and the current environment as temporally distinct views of the same world. 
The current environment reflects the latest application state, while memory may preserve earlier information that is no longer available there.
Consequently, a task intended to test memory may be solved by repeatedly searching the current device, while relying on memory alone may lead an agent to use outdated information. 
As illustrated in Figure~\ref{fig:recall_or_recheck}, an agent must therefore determine when to recall from memory, when to recheck the current environment, and when to combine both before acting.

\begin{table*}[t]
\centering
\tablestyle
\setlength{\tabcolsep}{5.0pt}
\caption{
Comparison of representative long-term memory and agent benchmarks.
\textbf{Temporal Separation} indicates that persistent memory and the
current environment correspond to different time points of the same world.
}
\label{tab:benchmark_comparison}

\resizebox{\textwidth}{!}{\begin{tabular}{@{}l ccc ccc c@{}}
\toprule
\multirow{2}{*}[-1.4ex]{\textbf{Benchmark}}
& \multicolumn{3}{c}{\textbf{Data Setting}}
& \multicolumn{3}{c}{\textbf{Agent Setting}}
& \multirow{2}{*}[-1.4ex]{\makecell{\textbf{Temporal}\\\textbf{Separation}}} \\
\cmidrule(lr){2-4}\cmidrule(lr){5-7}
& \makecell{\textbf{Long-Horizon}\\\textbf{History}}
& \textbf{Personalization}
& \textbf{Multimodal}
& \makecell{\textbf{Persistent}\\\textbf{Memory}}
& \makecell{\textbf{Environment}\\\textbf{Access}}
& \makecell{\textbf{Action}\\\textbf{Execution}}
& \\
\midrule

\tablegroup{8}{\textit{Long-term memory benchmarks}}
\addlinespace[1pt]

LoCoMo~\citep{maharana2024evaluating}
& \cmark & \xmark & \cmark & \cmark & \xmark & \xmark & \xmark \\

LongMemEval~\citep{wu2024longmemeval}
& \cmark & \xmark & \xmark & \cmark & \xmark & \xmark & \xmark \\

BEAM~\citep{tavakoli2026beyond}
& \cmark & \xmark & \xmark & \cmark & \xmark & \xmark & \xmark \\

\addlinespace[3pt]

\tablegroup{8}{\textit{Agent memory benchmarks}}
\addlinespace[1pt]

Mem2ActBench~\citep{shen2026mem2actbench}
& \cmark & \cmark & \xmark & \cmark & \xmark & \xmark & \xmark \\

MemoryArena~\citep{he2026memoryarena}
& \cmark & \xmark & \xmark & \cmark & \cmark & \cmark & \xmark \\

WorldMemArena~\citep{liu2026worldmemarena}
& \cmark & \xmark & \cmark & \cmark & \xmark & \xmark & \xmark \\

\addlinespace[3pt]

\tablegroup{8}{\textit{Mobile memory and agent benchmarks}}
\addlinespace[1pt]

MobileMem~\citep{deng2026mobilemem}
& \cmark & \cmark & \cmark & \cmark & \xmark & \xmark & \xmark \\

MemGUI-Bench~\citep{liu2026memgui}
& \cmark & \xmark & \cmark & \cmark & \cmark & \cmark & \xmark \\

KnowU-Bench~\citep{chen2026knowu}
& \cmark & \cmark & \cmark & \cmark & \cmark & \cmark & \xmark \\

\midrule

\rowcolor{PaperPanel}
\textbf{$\bm{R^2}$-MobileMemory}
& \cmark & \cmark & \cmark & \cmark & \cmark & \cmark & \cmark \\

\bottomrule
\end{tabular}}

\end{table*}

To study this problem, we introduce \textbf{$\bm{R^2}$-MobileMemory}, an executable benchmark that separates persistent memory $M_\tau$ from the current environment $E_T$ within the same evolving mobile world.
Memory is formed from observations available up to time $\tau$, whereas the environment continues to evolve until the evaluation time $T$.
Each task requires historical records retained in memory, current details on the device, or information from both sources.
Agents gather the needed information and interact with the phone through the GUI, and success is evaluated from the resulting application state.
We construct these worlds with \textbf{$\bm{R^2}$-Forge}, which synthesizes longitudinal user histories into executable application state and verifiable tasks.
Table~\ref{tab:benchmark_comparison} compares our setting with representative long-term memory and mobile-agent benchmarks.

Our contributions are threefold:

\begin{itemize}
    \item We introduce \textbf{$\bm{R^2}$-MobileMemory}, an executable benchmark of 513 tasks across ten year-long bilingual mobile worlds that explicitly separates persistent memory $M_\tau$ from the current environment $E_T$ to evaluate when agents should recall, recheck, or combine both sources.

    \item We develop \textbf{$\bm{R^2}$-Forge}, a construction pipeline that turns user profiles into evolving application records over more than a year and derives executable tasks from the resulting application state.

    \item Across three agents and eight memory backends, we show that persistent memory and App Search are complementary and that the main bottleneck is retrieval rather than storage. Most backends store over 90\% of the required earlier records, but agents recover only about half of them even on the strongest backends, and task difficulty grows with the number of required records rather than their age. We further find that simple dense retrieval matches more costly LLM-based backends and that forming memory up to evaluation time ($M_T$) trades historical tasks for current ones.
\end{itemize}

\section{Memory-Augmented Interaction in Evolving Mobile Environments}
\label{sec:problem}

Let $Q$ denote the space of user tasks, and let $E_t=(S_t,\mathcal{A})$ denote the mobile environment at time $t$, where $S_t$ is the observable application state and $\mathcal{A}$ is the set of executable GUI actions.
Unlike a static environment, $S_t$ evolves as application records are created, updated, or removed.

Let $M_\tau$ denote the persistent memory formed from observations available up to time $\tau$.
At evaluation time $T>\tau$, the agent interacts with $E_T$.
Memory $M_\tau$ may preserve historical records no longer exposed in $E_T$, whereas later updates in $E_T$ remain absent from $M_\tau$.

\begin{definition}[Memory-Augmented Mobile Agent Task]
\label{def:task}
Given a user task $q\in Q$, persistent memory $M_\tau$, and the current environment $E_T$, a memory-augmented mobile agent follows a policy $\pi$ to produce an interaction trajectory
\[
\xi=\pi(q,M_\tau,E_T).
\]
At each step, the agent may invoke memory search over $M_\tau$, search application records in $E_T$, or interact with $E_T$ through the GUI.
The objective is to gather the required information and use it to reach a terminal environment state satisfying the task specification.
\end{definition}

\section{\texorpdfstring{$\bm{R^2}$}{R2}-Forge: State-Grounded World-to-Task Construction}
\label{sec:r2forge}

\begin{figure*}[t]
\centering
\includegraphics[width=\textwidth]{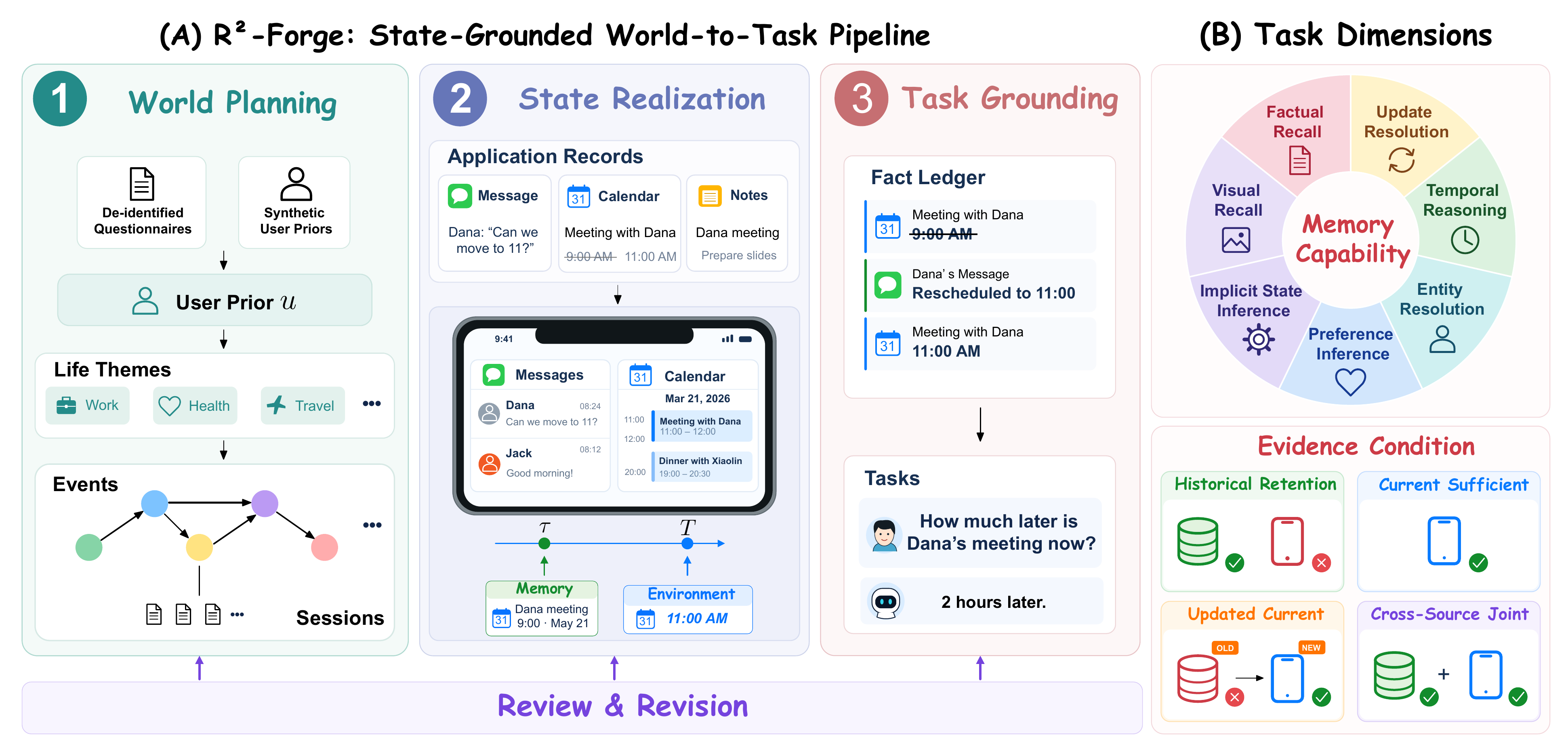}
\caption{
Overview of $R^2$-Forge and task dimensions.
(A) $R^2$-Forge plans a longitudinal user world, realizes it as executable
application state, and grounds tasks in the resulting records.
Persistent memory $M_\tau$ and the current environment $E_T$ are derived
from different temporal views of the same realized world, with review and
revision applied throughout construction.
(B) Tasks are organized by Memory Capability and Evidence Condition.
}
\label{fig:r2forge}
\end{figure*}

Implementing Definition~\ref{def:task} requires longitudinal mobile histories in which user behavior, application state, and multimodal records evolve over time. 
Collecting such histories at scale is costly and privacy-sensitive, while publicly available data provide limited coverage of sustained user-specific behavior~\citep{deng2026mobilemem,nie2026pspa,zhang2026lunar}.
Unconstrained synthesis, however, can deviate from realistic user behavior~\citep{xiao2026alpsbench,deng2026mobilemem}.
We therefore use controlled synthesis grounded in high-level behavioral priors.

As shown in Figure~\ref{fig:r2forge}, $R^2$-Forge constructs each world in three stages. \emph{World Planning} expands a user prior into a temporally ordered session history.
\emph{State Realization} instantiates these sessions as executable application records.
\emph{Task Grounding} derives verifiable tasks from the realized world. For the resulting application-record history $R$, a memory boundary $\tau$ and evaluation time $T$ define two temporal views of the same world:
\[
M_\tau = \mathrm{Ingest}(R_{\leq \tau}),
\qquad
E_T = \mathrm{Project}_T(R).
\]
Further details of benchmark construction and validation are provided in Appendix~\ref{app:construction}.

\subsection{Prior-Grounded World Planning}
\label{sec:world_planning}

Each world begins from a structured user prior $u$. Seven worlds are initialized from de-identified and fictionalized questionnaire priors describing smartphone habits and life patterns, while three use fully synthetic priors. 
All subsequent identities, events, application records, and media are generated within the simulated world.

$R^2$-Forge expands each prior into recurring life themes, concrete events, and temporally bounded sessions, producing an ordered session history
\[
H=(h_1,\ldots,h_n).
\]
Dependencies among events constrain their temporal ordering so that later activities can build on context established earlier.
To keep the application state changing throughout each timeline, every world also includes recurring storylines that revise the same contacts, schedules, notes, files, and settings several times over the year.
Once the timeline is fixed, we place a memory boundary $\tau$ between two sessions and choose a single evaluation time $T>\tau$ shared by all tasks in the world. Sessions up to $\tau$ provide the observations later used for memory formation, while subsequent sessions continue to evolve the world before $T$. 

\subsection{Application State Realization}
\label{sec:state_realization}

Given the planned session history $H$, $R^2$-Forge realizes each session as application-specific records under the guidance of Fact Intents and Task Blueprints. 
Fact Intents specify candidate facts and state transitions to be expressed in the world, while Task Blueprints reserve the capability and task coverage needed by downstream evaluation. 
Final supporting records and target outcomes are determined during task grounding.

Sessions are instantiated as heterogeneous application records, including messages, calendar events, notes, contacts, files, orders, and media.
Records referring to the same people, events, or application objects maintain consistent identities across sessions.
Application state persists across sessions, allowing later activities to update or remove objects created earlier.
Together, these records form the realized multimodal history $R$.

At evaluation time, each application exposes only its current state, so $E_T$ omits content that was edited or removed in earlier sessions. 
Memory $M_\tau$, formed from $R_{\leq\tau}$, may still contain this information.

\subsection{Task Grounding and Evidence Design}
\label{sec:task_construction}

After state realization, $R^2$-Forge extracts atomic facts from the realized record history $R$ into a Fact Ledger $F$. 
Each fact is linked to its source records and temporal context, allowing versions and relationships to be traced within the realized world. 
Task Blueprints are then resolved against $F$ to bind the required records and target outcome of each task while excluding unsupported planned content.
An LLM finally renders each grounded specification as a natural user request, avoiding rare proper nouns or verbatim strings from target records so that tasks cannot be solved through surface keyword matching alone.

We characterize each task along two orthogonal dimensions.
\emph{Memory Capability} describes the operation required over user-specific information, covering Factual Recall, Update Resolution, Temporal Reasoning, Entity Resolution, Preference Inference, Implicit State Inference, and Visual Recall.
\emph{Evidence Condition} describes how the required information is distributed between persistent memory and the current environment.
Historical Retention tasks require historical records that are no longer recoverable from $E_T$.
Cross-Source Joint tasks require both such historical records and current information introduced after $\tau$.
Updated Current tasks require a post-$\tau$ update that supersedes earlier information, and the current state alone is sufficient.
Current Sufficient tasks can be resolved entirely from $E_T$.
Panel~(B) of Figure~\ref{fig:r2forge} summarizes the two dimensions, with representative cases provided in Appendix~\ref{app:task_cases}.

\subsection{Review and Executable Validation}
\label{sec:validation}

Review and revision are interleaved throughout $R^2$-Forge.
Each final task is checked for three properties: its target is supported by the designated records, removing those records from $M_\tau$ or $E_T$ changes whether the task can be solved, and its outcome can be verified in the environment. These checks ensure that targets cannot be guessed from the request alone.
Detailed validation procedures are provided in Appendix~\ref{app:validation}.

Every task is executed in MobileGym~\citep{wu2026mobilegym} and evaluated from its terminal state.
A task-specific rubric specifies the required outcome, permitted changes,
and prohibited side effects.
Task success requires the terminal state to satisfy the complete rubric.

\paragraph{Benchmark statistics.}
$R^2$-MobileMemory contains 513 executable tasks across ten user worlds, including five Chinese (254 tasks) and five English (259 tasks) worlds.
The worlds span 462--547 days of simulated history with 87,459 application records, and the tasks cover
23 user applications, all seven Memory Capabilities, and all four Evidence Conditions.
Detailed world composition, temporal coverage, and task distributions are reported in Appendix~\ref{app:benchmark_details}.

\section{Experiments}
\label{sec:experiments}

Our experiments are designed to answer four questions.
\textbf{(RQ1)}~How do agents and memory backends compare in task success, and where is required information lost?
\textbf{(RQ2)}~How do persistent memory and App Search complement each other across Evidence Conditions?
\textbf{(RQ3)}~How do agents choose between memory and device search, and how does extending memory to evaluation time affect success?
\textbf{(RQ4)}~Is task difficulty driven by record age or record count, and does it stem from retrieval or downstream execution?

\subsection{Experimental Setup}
\label{sec:exp_setup}

We evaluate three multimodal mobile agents, UI-Venus-2-9B~\citep{team2026ui}, Gemini-3.7-Flash~\citep{gemini37flash}, and Qwen3.8-Max~\citep{qwen3.8}, paired with eight memory backends comprising NaiveRAG~\citep{lewis2020rag}, HippoRAG2~\citep{gutierrez2025rag}, A-MEM~\citep{xu2025amem}, MemOS~\citep{li2025memos}, EverMemOS~\citep{hu2026evermemos}, Mem0 and Mem0$^{g}$~\citep{chhikara2025mem0}, and LangMem~\citep{langchain2025langmem}.
Backends that require an LLM or an embedding model use Qwen3.6-27B~\citep{qwen36_27b} and Qwen3-Embedding-8B~\citep{qwen3embedding}, and each constructs its persistent memory $M_\tau$ from the application records available up to time $\tau$.

In the main setting, \emph{$M_\tau$ + App Search}, agents operate the phone through the GUI and can search both persistent memory $M_\tau$ and the application records in $E_T$, which we refer to as App Search.
To separate memory retrieval from downstream execution, we also evaluate an \emph{Oracle} setting that replaces memory search with the annotated supporting records of each task in the initial context, including earlier replaced versions without marking which version is valid.
Each task allows at most 50 interaction steps, and we report Task Success Rate, the fraction of tasks whose terminal state satisfies the complete rubric.
Full experimental details are provided in Appendix~\ref{app:experimental_setup}.

\subsection{Memory Backends and Coverage (RQ1)}
\label{sec:overall_results}

\begin{table*}[!t]
\centering
\tablestyle
\renewcommand{\arraystretch}{1.06}
\setlength{\tabcolsep}{2.6pt}
\caption{Task Success Rate (\%) under $M_\tau$ + App Search, broken down separately by Memory Capability and by Evidence Condition. \textit{No memory} removes memory and keeps App Search, and $\Delta$ is the overall gain over it. Memory Capabilities are Factual Recall (FR), Update Resolution (UR), Temporal Reasoning (TR), Entity Resolution (ER), Preference Inference (PI), Implicit State Inference (ISI), and Visual Recall (VR). Evidence Conditions are Historical Retention (HR), Cross-Source Joint (CSJ), Updated Current (UC), and Current Sufficient (CS). For each agent, \textbf{bold} marks the best backend and \underline{underline} marks the second best.}
\label{tab:main_results}
\label{tab:evidence_condition_results}

\begin{tabular*}{\textwidth}{@{\extracolsep{\fill}}l cc @{\extracolsep{\fill}\hspace{1.1em}} ccccccc @{\extracolsep{\fill}\hspace{1.1em}} cccc@{}}
\toprule
\multirow{2}{*}[-0.5ex]{\textbf{Memory Backend}} & \multirow{2}{*}[-0.5ex]{\textbf{Overall}} & \multirow{2}{*}[-0.5ex]{\textbf{$\bm{\Delta}$}}
& \multicolumn{7}{c}{\textbf{By Memory Capability}}
& \multicolumn{4}{c}{\textbf{By Evidence Condition}} \\
\cmidrule(lr){4-10} \cmidrule(l){11-14}
& &
& \textbf{FR} & \textbf{UR} & \textbf{TR} & \textbf{ER} & \textbf{PI} & \textbf{ISI} & \textbf{VR} & \textbf{HR} & \textbf{CSJ} & \textbf{UC} & \textbf{CS} \\
\midrule
\tablegroup{14}{UI-Venus-2-9B}
\rowcolor{TableSubtle}
\textit{No memory} & 5.1 & --- & 11.8 & 2.4 & 1.2 & 2.5 & 9.4 & 4.6 & 4.9 & 0.0 & 1.2 & 8.5 & 12.8 \\
\cmidrule(lr){1-14}
NaiveRAG & 5.8 & +0.8 & \underline{18.4} & \underline{1.2} & 0.0 & 1.2 & 4.7 & 9.2 & \textbf{8.2} & \underline{4.7} & \underline{2.5} & 4.7 & 12.8 \\
HippoRAG2 & 6.4 & +1.4 & 14.5 & \underline{1.2} & 0.0 & \textbf{4.9} & \textbf{9.4} & 9.2 & \textbf{8.2} & 3.9 & 1.9 & \underline{11.3} & 11.1 \\
A-MEM & 5.1 & +0.0 & 13.2 & \underline{1.2} & \underline{1.2} & 1.2 & 3.1 & \underline{10.8} & \underline{6.6} & 3.9 & \textbf{3.1} & 6.6 & 7.7 \\
Mem0 & \underline{6.8} & \underline{+1.8} & \textbf{19.7} & \textbf{2.4} & \underline{1.2} & \underline{3.7} & \textbf{9.4} & 4.6 & \textbf{8.2} & 2.3 & 0.6 & \underline{11.3} & \textbf{16.2} \\
Mem0$^{g}$ & 5.3 & +0.2 & \underline{18.4} & \underline{1.2} & \textbf{2.4} & 2.5 & 6.2 & 4.6 & 1.6 & 0.8 & 0.6 & 6.6 & \underline{15.4} \\
MemOS & 3.3 & $-$1.8 & 7.9 & \underline{1.2} & 0.0 & 1.2 & 4.7 & 4.6 & 4.9 & 0.8 & 0.0 & 9.4 & 5.1 \\
EverMemOS & \textbf{7.0} & \textbf{+1.9} & 17.1 & \underline{1.2} & \underline{1.2} & \textbf{4.9} & \underline{7.8} & \textbf{12.3} & \underline{6.6} & \textbf{5.5} & 1.2 & \textbf{12.3} & 12.0 \\
LangMem & 4.5 & $-$0.6 & 15.8 & \underline{1.2} & 0.0 & 1.2 & 6.2 & 4.6 & 3.3 & 3.1 & 0.6 & 5.7 & 10.3 \\
\addlinespace[3pt]
\tablegroup{14}{Gemini-3.7-Flash}
\rowcolor{TableSubtle}
\textit{No memory} & 17.7 & --- & 32.9 & 9.6 & 16.9 & 6.2 & 18.8 & 20.0 & 23.0 & 1.6 & 2.5 & 34.9 & 41.0 \\
\cmidrule(lr){1-14}
NaiveRAG & \underline{32.6} & \underline{+14.8} & 46.1 & 9.6 & \underline{25.3} & \textbf{23.5} & \underline{43.8} & \underline{35.4} & 54.1 & \underline{30.5} & \underline{18.5} & \underline{43.4} & 44.4 \\
HippoRAG2 & 25.5 & +7.8 & 36.8 & 7.2 & 14.5 & 18.5 & 29.7 & 30.8 & 50.8 & 23.4 & 13.6 & 34.9 & 35.9 \\
A-MEM & 31.8 & +14.0 & 46.1 & \underline{10.8} & \underline{25.3} & 14.8 & 39.1 & \textbf{38.5} & \textbf{59.0} & \textbf{31.2} & 17.3 & 39.6 & \underline{45.3} \\
Mem0 & 18.5 & +0.8 & 32.9 & 4.8 & 14.5 & 16.0 & 14.1 & 24.6 & 26.2 & 7.8 & 7.4 & 30.2 & 35.0 \\
Mem0$^{g}$ & 24.2 & +6.4 & 40.8 & \textbf{12.0} & 16.9 & 14.8 & 28.1 & 27.7 & 34.4 & 8.6 & 9.3 & 40.6 & \textbf{47.0} \\
MemOS & 23.2 & +5.5 & 42.1 & 7.2 & 15.7 & 8.6 & 32.8 & 26.2 & 37.7 & 7.0 & 7.4 & \underline{43.4} & 44.4 \\
EverMemOS & \textbf{32.7} & \textbf{+15.0} & \underline{47.4} & 9.6 & \textbf{26.5} & \underline{19.8} & \textbf{45.3} & \underline{35.4} & \underline{55.7} & 25.8 & \textbf{26.5} & \underline{43.4} & 39.3 \\
LangMem & 27.5 & +9.7 & \textbf{48.7} & 8.4 & \underline{25.3} & 18.5 & 25.0 & 30.8 & 41.0 & 15.6 & 13.0 & \textbf{45.3} & 44.4 \\
\addlinespace[3pt]
\tablegroup{14}{Qwen3.8-Max}
\rowcolor{TableSubtle}
\textit{No memory} & 29.2 & --- & 39.5 & 19.3 & 25.3 & 25.9 & 31.2 & 35.4 & 31.1 & 0.0 & 0.0 & 60.4 & 73.5 \\
\cmidrule(lr){1-14}
NaiveRAG & \underline{56.3} & \underline{+27.1} & \textbf{69.7} & 33.7 & \textbf{44.6} & \underline{45.7} & \underline{65.6} & 64.6 & \textbf{82.0} & \underline{48.4} & \underline{38.9} & \textbf{71.7} & \underline{75.2} \\
HippoRAG2 & 53.4 & +24.2 & 61.8 & \underline{38.6} & \underline{41.0} & \textbf{49.4} & 57.8 & 63.1 & 70.5 & 47.7 & 35.8 & 64.2 & 74.4 \\
A-MEM & \textbf{57.5} & \textbf{+28.3} & \underline{67.1} & \textbf{39.8} & \underline{41.0} & \textbf{49.4} & \textbf{70.3} & \textbf{69.2} & \underline{77.0} & \textbf{51.6} & \textbf{43.2} & 66.0 & \textbf{76.1} \\
Mem0 & 37.4 & +8.2 & 50.0 & 18.1 & 26.5 & 39.5 & 37.5 & 43.1 & 54.1 & 13.3 & 14.8 & 64.2 & 70.9 \\
Mem0$^{g}$ & 38.2 & +9.0 & 47.4 & 24.1 & 25.3 & 37.0 & 39.1 & 47.7 & 54.1 & 10.9 & 19.1 & \underline{68.9} & 66.7 \\
MemOS & 36.3 & +7.0 & 46.1 & 19.3 & 24.1 & 33.3 & 43.8 & 46.2 & 49.2 & 14.1 & 13.6 & 63.2 & 67.5 \\
EverMemOS & 48.9 & +19.7 & \underline{67.1} & 22.9 & 32.5 & 40.7 & 60.9 & \underline{66.2} & 63.9 & 36.7 & 31.5 & 64.2 & 72.6 \\
LangMem & 40.4 & +11.1 & 52.6 & 22.9 & 34.9 & 35.8 & 39.1 & 47.7 & 55.7 & 16.4 & 19.8 & 65.1 & 72.6 \\
\bottomrule
\end{tabular*}
\end{table*}

\paragraph{Simple dense retrieval matches LLM-based memory backends.}
In Table~\ref{tab:main_results}, NaiveRAG performs on par with the strongest backend on each capable agent, reaching 56.3\% against 57.5\% for A-MEM on Qwen3.8-Max and 32.6\% against 32.7\% for EverMemOS on Gemini-3.7-Flash.
On Qwen3.8-Max, Mem0, Mem0$^{g}$, MemOS, and LangMem trail A-MEM by 17--21 points.
These differences come mainly from Historical Retention and Cross-Source Joint, where success on Qwen3.8-Max ranges from 10.9\% to 51.6\% and from 13.6\% to 43.2\% across backends.
On Current Sufficient, success stays within 66.7--76.1\% because App Search can recover information that remains on the device.
Agent capability matters more than backend choice, as UI-Venus-2-9B remains between 3.3\% and 7.0\% across all backends.

\begin{figure}[!t]
\centering
\includegraphics[width=\textwidth]{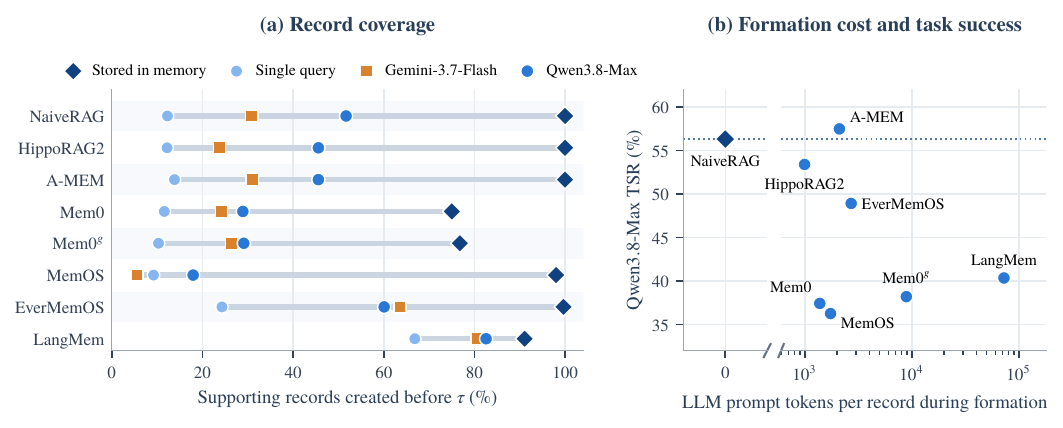}
\caption{\textbf{Record coverage and memory formation cost.}
(a) On Historical Retention and Cross-Source Joint tasks, the share of supporting records created before $\tau$ that are stored in memory, returned among the top ten results of a single query with the task request, and retrieved by each agent during execution.
(b) LLM prompt tokens per record during memory formation versus Task Success Rate on Qwen3.8-Max.}
\label{fig:coverage_cost}
\end{figure}

\paragraph{Information is lost mainly at retrieval rather than storage.}
Figure~\ref{fig:coverage_cost}a measures the share of pre-$\tau$ supporting records on Historical Retention and Cross-Source Joint tasks that are stored in $M_\tau$, returned by a single top-10 query, and retrieved by the agent during execution.
Six of the eight backends retain over 90\% of these records, while Mem0 and Mem0$^{g}$ lose about a quarter during fact extraction.
In contrast, a single query returns fewer than 15\% of the required records for most backends, and even with iterative search, agents recover only about half on the strongest backends.
High storage retention therefore does not guarantee retrieval, as MemOS stores 98\% of the required records yet Qwen3.8-Max retrieves only 18\%.

\paragraph{Higher formation cost does not improve task success.}
Figure~\ref{fig:coverage_cost}b compares LLM prompt tokens per record during memory formation against task success on Qwen3.8-Max.
NaiveRAG uses no LLM calls during formation yet matches A-MEM, whereas other backends spend from one thousand to over seventy thousand prompt tokens per record without higher task success.
LangMem incurs the highest formation cost by merging records into a few long documents. This yields the highest coverage in Figure~\ref{fig:coverage_cost}a, yet each retrieved entry is a long document and task success reaches only 40.4\%.

\subsection{Memory and App Search (RQ2)}
\label{sec:access_ablation}

\paragraph{Persistent memory and App Search are complementary.}
Table~\ref{tab:access_ablation} ablates persistent memory $M_\tau$ and App Search from the main setting using NaiveRAG as the memory backend.
With GUI interaction alone, no agent exceeds 4\% task success.
Adding App Search raises Qwen3.8-Max to 29.2\% by solving Current Sufficient at 73.5\% and Updated Current at 60.4\%, but leaves Historical Retention and Cross-Source Joint at 0.0\% because pre-$\tau$ records have disappeared from the device.
Conversely, memory without App Search reaches 32.0\% on Historical Retention but only 9.4\% on Updated Current, where post-$\tau$ updates must be located on the phone.
Combining both sources raises overall success to 56.3\% and lifts Cross-Source Joint from at most 10.5\% with either tool to 38.9\%.

\begin{table}[!t]
\centering
\tablestyle
\setlength{\tabcolsep}{9pt}
\caption{Task Success Rate (\%) when App Search or memory is removed. All settings include GUI interaction, and $M_\tau$ uses NaiveRAG. Bold marks the best result per agent.}
\label{tab:access_ablation}

\begin{tabular}{@{}ccc cccc@{}}
\toprule
\multirow{2}{*}[-1.6ex]{\textbf{App Search}} & \multirow{2}{*}[-1.6ex]{\textbf{$\bm{M_\tau}$}}
& \multirow{2}{*}[-1.6ex]{\textbf{Overall}}
& \multicolumn{2}{c}{\textbf{Requires Pre-$\bm{\tau}$ Records}}
& \multicolumn{2}{c}{\textbf{Available in $\bm{E_T}$}} \\
\cmidrule(lr){4-5} \cmidrule(lr){6-7}
& &
& \makecell{\textbf{Historical}\\\textbf{Retention}}
& \makecell{\textbf{Cross-Source}\\\textbf{Joint}}
& \makecell{\textbf{Updated}\\\textbf{Current}}
& \makecell{\textbf{Current}\\\textbf{Sufficient}} \\
\midrule
\tablegroup{7}{UI-Venus-2-9B}
\xmark & \xmark & 0.4 & 0.0 & 0.0 & 0.9 & 0.9 \\
\cmark & \xmark & 5.1 & 0.0 & 1.2 & \textbf{8.5} & \textbf{12.8} \\
\xmark & \cmark & 2.7 & \textbf{4.7} & 0.6 & 0.0 & 6.0 \\
\rowcolor{TableSubtle}
\cmark & \cmark & \textbf{5.8} & \textbf{4.7} & \textbf{2.5} & 4.7 & \textbf{12.8} \\
\addlinespace[3pt]
\tablegroup{7}{Gemini-3.7-Flash}
\xmark & \xmark & 1.8 & 0.0 & 0.6 & 0.9 & 6.0 \\
\cmark & \xmark & 17.7 & 1.6 & 2.5 & 34.9 & 41.0 \\
\xmark & \cmark & 9.0 & 24.2 & 1.9 & 0.0 & 10.3 \\
\rowcolor{TableSubtle}
\cmark & \cmark & \textbf{32.6} & \textbf{30.5} & \textbf{18.5} & \textbf{43.4} & \textbf{44.4} \\
\addlinespace[3pt]
\tablegroup{7}{Qwen3.8-Max}
\xmark & \xmark & 3.9 & 0.0 & 0.6 & 5.7 & 11.1 \\
\cmark & \xmark & 29.2 & 0.0 & 0.0 & 60.4 & 73.5 \\
\xmark & \cmark & 18.1 & 32.0 & 10.5 & 9.4 & 21.4 \\
\rowcolor{TableSubtle}
\cmark & \cmark & \textbf{56.3} & \textbf{48.4} & \textbf{38.9} & \textbf{71.7} & \textbf{75.2} \\
\bottomrule
\end{tabular}
\end{table}

\subsection{Tool Use and Memory Formation Time (RQ3)}
\label{sec:recall_or_recheck}

Because agents are not told which Evidence Condition a task belongs to, they must decide from the request whether to recall from memory, recheck the device, or combine both.

\begin{figure}[!t]
\centering
\includegraphics[width=\textwidth]{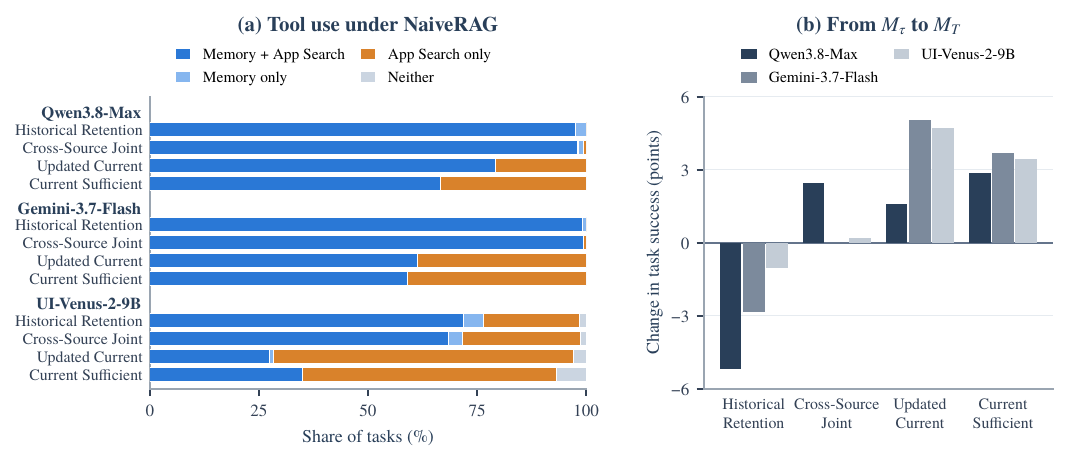}
\caption{\textbf{Tool use and memory formed up to $T$.}
(a) Share of tasks in which each agent queries memory, App Search, or both under NaiveRAG, by Evidence Condition.
(b) Change in task success from $M_\tau$ to $M_T$ by Evidence Condition, averaged over NaiveRAG, A-MEM, and EverMemOS.}
\label{fig:version_choice_mt}
\end{figure}

\paragraph{Strong agents adapt their tool use to the task.}
Figure~\ref{fig:version_choice_mt}a reports the share of tasks in which each agent queries memory, App Search, or both under NaiveRAG.
On Historical Retention and Cross-Source Joint tasks, Qwen3.8-Max and Gemini-3.7-Flash query both sources in 98--99\% of tasks, whereas on Updated Current and Current Sufficient tasks they skip memory and rely on App Search alone in 21--41\% of tasks.
Their failures on Historical Retention tasks thus stem mainly from incomplete retrieval rather than failing to query memory.
In contrast, UI-Venus-2-9B uses App Search alone in 22\% of Historical Retention and 27\% of Cross-Source Joint tasks, failing every such run.

\paragraph{Extending memory to evaluation time trades off historical and current tasks.}
Figure~\ref{fig:version_choice_mt}b compares $M_\tau$ with memory formed from all records up to $T$, denoted $M_T$, averaged over NaiveRAG, A-MEM, and EverMemOS.
Across all three agents, extending memory to $T$ improves success on Updated Current and Current Sufficient tasks by incorporating recent updates, but reduces success on Historical Retention tasks because later records compete with earlier ones.
Over all tasks, these gains and losses largely cancel, changing overall success by less than three points for every agent and backend.

\subsection{Drivers of Task Difficulty (RQ4)}
\label{sec:evidence_chain}

\begin{figure}[!t]
\centering
\includegraphics[width=\textwidth]{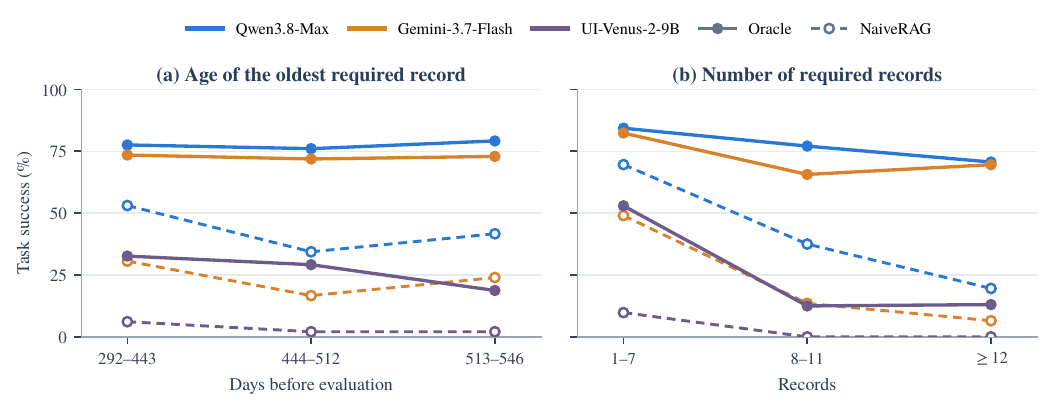}
\caption{\textbf{Task success by the age and number of required records.}
Historical Retention and Cross-Source Joint tasks are divided into three equal-sized groups by (a) the age of the oldest required record at evaluation time and (b) the number of required records.
Solid lines show the Oracle setting and dashed lines show NaiveRAG under $M_\tau$ + App Search.}
\label{fig:rq4_difficulty}
\end{figure}

Over all 513 tasks, the Oracle setting raises task success to 28.1\% on UI-Venus-2-9B, 70.0\% on Gemini-3.7-Flash, and 79.3\% on Qwen3.8-Max, compared with 5.8\%, 32.6\%, and 56.3\% under NaiveRAG.
Because Updated Current and Current Sufficient tasks can be solved directly from the device via App Search, we focus on Historical Retention and Cross-Source Joint tasks, where completion requires pre-$\tau$ records from memory.
Figure~\ref{fig:rq4_difficulty} compares NaiveRAG under $M_\tau$ + App Search with the Oracle setting on these tasks across three equal-sized groups of record age and required record count.

\paragraph{Difficulty scales with record count rather than record age.}
In Figure~\ref{fig:rq4_difficulty}a, across the three age groups spanning 292 to 546 days before evaluation, Oracle success on Qwen3.8-Max and Gemini-3.7-Flash varies by less than four points, and NaiveRAG success does not decline monotonically with age.
In Figure~\ref{fig:rq4_difficulty}b, as the number of required records grows from 1--7 to at least 12, NaiveRAG success drops from 69.6\% to 19.6\% on Qwen3.8-Max and from 49.0\% to 6.5\% on Gemini-3.7-Flash.
Under the Oracle, this decline shrinks to 13.6 and 12.8 points. Across all Historical Retention and Cross-Source Joint tasks, the gap between the two agents also narrows from 19.3 points to 4.8 points.
For strong agents, task difficulty therefore lies primarily in retrieving complete multi-record sets rather than reasoning over them, whereas UI-Venus-2-9B still drops from 52.9\% to 13.0\% under the Oracle due to downstream execution limits.

\section{Related Work}

\paragraph{Long-term memory systems and evaluation.}
Long-term memory systems organize past experience through context management~\citep{packer2023memgpt}, fact extraction and consolidation~\citep{chhikara2025mem0,chen2025telemem}, hierarchical and episodic memory stores~\citep{li2025memos,hu2026evermemos}, linked memories~\citep{xu2025amem}, and associative retrieval~\citep{gutierrez2025rag}.
Memory evaluation has expanded from long-history recall and reasoning~\citep{maharana2024evaluating,wu2024longmemeval,tavakoli2026beyond} to incremental memory management, failure diagnosis~\citep{hu2026evaluating,chen2025halumem,deng2026memtrace}, and memory-supported agent behavior~\citep{shen2026mem2actbench,he2026memoryarena,liu2026worldmemarena}.
These works establish that memory quality depends on both what is retained and how it is retrieved and used.
Our evaluation extends this perspective to settings where persistent memory and the current environment provide distinct, temporally separated sources of information.

\paragraph{Personalized and executable mobile agents.}
Recent benchmarks study personalization from behavioral histories~\citep{chen2026knowu,nie2026pspa} and evaluate agents through stateful mobile interaction~\citep{kong2026mobileworld,liu2026memgui,zhu2026mobilepa}.
MobileMem~\citep{deng2026mobilemem} evaluates longitudinal mobile memory from accumulated experience, while MobilePA-Bench~\citep{zhu2026mobilepa} combines persistent-memory access with stateful tools.
\textbf{$\bm{R^2}$-MobileMemory} differs by explicitly controlling when memory is formed relative to the current world and grounding tasks according to whether the required information resides in memory, the current environment, or both.

\section{Conclusion}

We introduced \textbf{$\bm{R^2}$-MobileMemory} to evaluate memory-augmented agents when persistent memory and the current mobile environment provide different temporal views of the same evolving world.
Our results place the main bottleneck at retrieval rather than storage.
Memory and App Search are complementary, and strong agents query memory when tasks need earlier records.
Yet agents recover only about half of the earlier records they need, even on the strongest backends, and success drops as the number of required records grows.
Simple dense retrieval matches more costly LLM-based backends, and forming memory up to evaluation time trades historical tasks for current ones.
Future memory-augmented agents therefore need to decide when to recall or recheck, retrieve complete multi-record evidence, and incorporate updates without displacing earlier versions.

\bibliography{references}

\begin{thebibliography}{41}
\providecommand{\natexlab}[1]{#1}
\providecommand{\url}[1]{\texttt{#1}}
\expandafter\ifx\csname urlstyle\endcsname\relax
  \providecommand{\doi}[1]{doi: #1}\else
  \providecommand{\doi}{doi: \begingroup \urlstyle{rm}\Url}\fi

\bibitem[Bei et~al.(2026)Bei, Wei, Ning, Zhao, Liu, Lin, Zhu, Hamann, He, and
  Tong]{bei2026mem}
Yuanchen Bei, Tianxin Wei, Xuying Ning, Yanjun Zhao, Zhining Liu, Xiao Lin,
  Yada Zhu, Hendrik Hamann, Jingrui He, and Hanghang Tong.
\newblock Mem-gallery: Benchmarking multimodal long-term conversational memory
  for mllm agents.
\newblock In \emph{Proceedings of the 64th Annual Meeting of the Association
  for Computational Linguistics (Volume 1: Long Papers)}, pages 40750--40784,
  2026.

\bibitem[Chai et~al.(2026)Chai, Wang, Yang, Peng, Song, Fu, Liu, Lin, Wang, and
  Zhang]{chai2026smmbench}
Huacan Chai, Yukai Wang, Yingxuan Yang, Dan Peng, Yuanyi Song, Zhihui Fu,
  Weiwen Liu, Jianghao Lin, Jun Wang, and Weinan Zhang.
\newblock Smmbench: A benchmark for source-distributed multimodal agent memory.
\newblock \emph{arXiv preprint arXiv:2605.15710}, 2026.

\bibitem[Chen et~al.(2025{\natexlab{a}})Chen, Guan, Lin, Li, Lin, Wang, Chen,
  Luo, Sun, Zhang, et~al.]{chen2025telemem}
Chunliang Chen, Ming Guan, Xiao Lin, Jiaxu Li, Luxi Lin, Qiyi Wang, Xiangyu
  Chen, Jixiang Luo, Changzhi Sun, Dell Zhang, et~al.
\newblock Telemem: Building long-term and multimodal memory for agentic ai.
\newblock \emph{arXiv preprint arXiv:2601.06037}, 2025{\natexlab{a}}.

\bibitem[Chen et~al.(2025{\natexlab{b}})Chen, Niu, Li, Liu, Zheng, Tang, Li,
  Xiong, and Li]{chen2025halumem}
Ding Chen, Simin Niu, Kehang Li, Peng Liu, Xiangping Zheng, Bo~Tang, Xinchi Li,
  Feiyu Xiong, and Zhiyu Li.
\newblock Halumem: Evaluating hallucinations in memory systems of agents.
\newblock \emph{arXiv preprint arXiv:2511.03506}, 2025{\natexlab{b}}.

\bibitem[Chen et~al.(2026)Chen, Lu, Xu, Shao, Zhao, Tang, Du, Song, Liu, Yan,
  et~al.]{chen2026knowu}
Tongbo Chen, Zhengxi Lu, Zhan Xu, Guocheng Shao, Shaohan Zhao, Fei Tang, Yong
  Du, Kaitao Song, Yizhou Liu, Yuchen Yan, et~al.
\newblock Knowu-bench: Towards interactive, proactive, and personalized mobile
  agent evaluation.
\newblock \emph{arXiv preprint arXiv:2604.08455}, 2026.

\bibitem[Chhikara et~al.(2025)Chhikara, Khant, Aryan, Singh, and
  Yadav]{chhikara2025mem0}
Prateek Chhikara, Dev Khant, Saket Aryan, Taranjeet Singh, and Deshraj Yadav.
\newblock Mem0: Building production-ready ai agents with scalable long-term
  memory.
\newblock \emph{arXiv preprint arXiv:2504.19413}, 2025.

\bibitem[Deng et~al.(2026{\natexlab{a}})Deng, Xue, Ru, Xu, Qiao, Wang, Chen,
  Xu, Jiang, Jiang, et~al.]{deng2026mobilemem}
Xinle Deng, Yida Xue, Xiangyuan Ru, Haoming Xu, Shuofei Qiao, Mengru Wang,
  Yijun Chen, Buqiang Xu, Chen Jiang, Yuchen~Eleanor Jiang, et~al.
\newblock Mobilemem: Learning from a year of mobile experiences.
\newblock \emph{arXiv preprint arXiv:2608.13606}, 2026{\natexlab{a}}.

\bibitem[Deng et~al.(2026{\natexlab{b}})Deng, Zhong, Peng, Lu, Wu, Li, Xu, Yao,
  Fang, Cao, et~al.]{deng2026memtrace}
Xinle Deng, Ruobin Zhong, Hujin Peng, Xiaoben Lu, Yanzhe Wu, Guang Li, Buqiang
  Xu, Yunzhi Yao, Jizhan Fang, Haoliang Cao, et~al.
\newblock Memtrace: Tracing and attributing errors in large language model
  memory systems.
\newblock \emph{arXiv preprint arXiv:2605.28732}, 2026{\natexlab{b}}.

\bibitem[{Google DeepMind}(2026)]{gemini37flash}
{Google DeepMind}.
\newblock Gemini 3.7 flash model card, August 2026.
\newblock URL
  \url{https://deepmind.google/models/model-cards/gemini-3-7-flash/}.

\bibitem[Guti{\'e}rrez et~al.(2025)Guti{\'e}rrez, Shu, Qi, Zhou, and
  Su]{gutierrez2025rag}
Bernal~Jim{\'e}nez Guti{\'e}rrez, Yiheng Shu, Weijian Qi, Sizhe Zhou, and
  Yu~Su.
\newblock From rag to memory: Non-parametric continual learning for large
  language models.
\newblock \emph{arXiv preprint arXiv:2502.14802}, 2025.

\bibitem[He et~al.(2026)He, Wang, Zhi, Hu, Chen, Yin, Chen, Wu, Ouyang, Wang,
  et~al.]{he2026memoryarena}
Zexue He, Yu~Wang, Churan Zhi, Yuanzhe Hu, Tzu-Ping Chen, Lang Yin, Ze~Chen,
  Tong~Arthur Wu, Siru Ouyang, Zihan Wang, et~al.
\newblock Memoryarena: Benchmarking agent memory in interdependent
  multi-session agentic tasks.
\newblock \emph{arXiv preprint arXiv:2602.16313}, 2026.

\bibitem[Hu et~al.(2026{\natexlab{a}})Hu, Gao, Zhou, Xu, Bai, Li, Zhang, Li,
  Zhang, Bing, and Deng]{hu2026evermemos}
Chuanrui Hu, Xingze Gao, Zuyi Zhou, Dannong Xu, Yi~Bai, Xintong Li, Hui Zhang,
  Tong Li, Chong Zhang, Lidong Bing, and Yafeng Deng.
\newblock {EverMemOS}: A self-organizing memory operating system for structured
  long-horizon reasoning.
\newblock \emph{arXiv preprint arXiv:2601.02163}, 2026{\natexlab{a}}.
\newblock URL \url{https://arxiv.org/abs/2601.02163}.

\bibitem[Hu et~al.(2026{\natexlab{b}})Hu, Wang, and McAuley]{hu2026evaluating}
Yuanzhe Hu, Yu~Wang, and Julian McAuley.
\newblock Evaluating memory in llm agents via incremental multi-turn
  interactions.
\newblock In \emph{International Conference on Learning Representations},
  volume 2026, pages 156259--156291, 2026{\natexlab{b}}.

\bibitem[Hu et~al.(2025)Hu, Liu, Yue, Zhang, Liu, Zhu, Lin, Guo, Dou, Xi,
  et~al.]{hu2025memory}
Yuyang Hu, Shichun Liu, Yanwei Yue, Guibin Zhang, Boyang Liu, Fangyi Zhu,
  Jiahang Lin, Honglin Guo, Shihan Dou, Zhiheng Xi, et~al.
\newblock Memory in the age of ai agents.
\newblock \emph{arXiv preprint arXiv:2512.13564}, 2025.

\bibitem[Kong et~al.(2026)Kong, Zhang, Yang, Gao, Liu, Tong, Cai, Zhou, Zhang,
  Chen, et~al.]{kong2026mobileworld}
Quyu Kong, Xu~Zhang, Zhenyu Yang, Nolan Gao, Chen Liu, Panrong Tong, Chenglin
  Cai, Hanzhang Zhou, Jianan Zhang, Liangyu Chen, et~al.
\newblock Mobileworld: Benchmarking autonomous mobile agents in agent-user
  interactive and mcp-augmented environments.
\newblock In \emph{Proceedings of the 64th Annual Meeting of the Association
  for Computational Linguistics (Volume 1: Long Papers)}, pages 6142--6167,
  2026.

\bibitem[{LangChain Team}(2025)]{langchain2025langmem}
{LangChain Team}.
\newblock {LangMem SDK} for agent long-term memory.
\newblock Official documentation and release announcement, 2025.
\newblock URL \url{https://www.langchain.com/blog/langmem-sdk-launch}.

\bibitem[Lewis et~al.(2020)Lewis, Perez, Piktus, Petroni, Karpukhin, Goyal,
  K{\"u}ttler, Lewis, Yih, Rockt{\"a}schel, Riedel, and Kiela]{lewis2020rag}
Patrick Lewis, Ethan Perez, Aleksandra Piktus, Fabio Petroni, Vladimir
  Karpukhin, Naman Goyal, Heinrich K{\"u}ttler, Mike Lewis, Wen-tau Yih, Tim
  Rockt{\"a}schel, Sebastian Riedel, and Douwe Kiela.
\newblock Retrieval-augmented generation for knowledge-intensive {NLP} tasks.
\newblock In \emph{Advances in Neural Information Processing Systems},
  volume~33, 2020.
\newblock URL \url{https://arxiv.org/abs/2005.11401}.

\bibitem[Li et~al.(2017)Li, Su, Shen, Li, Cao, and Niu]{li2017dailydialog}
Yanran Li, Hui Su, Xiaoyu Shen, Wenjie Li, Ziqiang Cao, and Shuzi Niu.
\newblock Dailydialog: A manually labelled multi-turn dialogue dataset.
\newblock In \emph{Proceedings of the Eighth International Joint Conference on
  Natural Language Processing (Volume 1: Long Papers)}, pages 986--995, 2017.

\bibitem[Li et~al.(2025)Li, Song, Xi, Wang, et~al.]{li2025memos}
Zhiyu Li, Shichao Song, Chenyang Xi, Hanyu Wang, et~al.
\newblock {MemOS}: A memory {OS} for {AI} system.
\newblock \emph{arXiv preprint arXiv:2507.03724}, 2025.
\newblock URL \url{https://arxiv.org/abs/2507.03724}.

\bibitem[Liu et~al.(2026{\natexlab{a}})Liu, Yang, Pu, Liu, Long, Guo, Chen,
  Weng, Kochkina, Kaur, et~al.]{liu2026worldmemarena}
Chengzhi Liu, Yuzhe Yang, Sophia~Xiao Pu, Yepeng Liu, Lin Long, Yichen Guo, Nuo
  Chen, Zhaotian Weng, Elena Kochkina, Simerjot Kaur, et~al.
\newblock Worldmemarena: Evaluating multimodal agent memory through
  action-world interaction.
\newblock \emph{arXiv preprint arXiv:2605.29341}, 2026{\natexlab{a}}.

\bibitem[Liu et~al.(2026{\natexlab{b}})Liu, Zhao, Liang, Luo, Tang, Chai, Lin,
  Xiao, Wang, Chen, et~al.]{liu2026memgui}
Guangyi Liu, Pengxiang Zhao, Yaozhen Liang, Qinyi Luo, Shunye Tang, Yuxiang
  Chai, Weifeng Lin, Han Xiao, WenHao Wang, Siheng Chen, et~al.
\newblock Memgui-bench: Benchmarking memory of mobile gui agents in dynamic
  environments.
\newblock \emph{arXiv preprint arXiv:2602.06075}, 2026{\natexlab{b}}.

\bibitem[Maharana et~al.(2024)Maharana, Lee, Tulyakov, Bansal, Barbieri, and
  Fang]{maharana2024evaluating}
Adyasha Maharana, Dong-Ho Lee, Sergey Tulyakov, Mohit Bansal, Francesco
  Barbieri, and Yuwei Fang.
\newblock Evaluating very long-term conversational memory of llm agents.
\newblock In \emph{Proceedings of the 62nd Annual Meeting of the Association
  for Computational Linguistics (Volume 1: Long Papers)}, pages 13851--13870,
  2024.

\bibitem[Nie et~al.(2026)Nie, Liu, Bai, Wang, Liu, Yao, and Wang]{nie2026pspa}
Hongyi Nie, Xunyuan Liu, Yudong Bai, Yaqing Wang, Yang Liu, Quanming Yao, and
  Zhen Wang.
\newblock Pspa-bench: A personalized benchmark for smartphone gui agent.
\newblock \emph{arXiv preprint arXiv:2603.29318}, 2026.

\bibitem[Packer et~al.(2023)Packer, Wooders, Lin, Fang, Patil, Stoica, and
  Gonzalez]{packer2023memgpt}
Charles Packer, Sarah Wooders, Kevin Lin, Vivian Fang, Shishir~G Patil, Ion
  Stoica, and Joseph~E Gonzalez.
\newblock Memgpt: Towards llms as operating systems.
\newblock \emph{arXiv preprint arXiv:2310.08560}, 2023.

\bibitem[{Qwen Team}(2026{\natexlab{a}})]{qwen3.8}
{Qwen Team}.
\newblock {Qwen3.8-Max}: A new bar for coding and cowork, August
  2026{\natexlab{a}}.
\newblock URL \url{https://qwen.ai/blog?id=qwen3.8}.

\bibitem[{Qwen Team}(2026{\natexlab{b}})]{qwen36_27b}
{Qwen Team}.
\newblock {Qwen3.6-27B}: Flagship-level coding in a 27b dense model, April
  2026{\natexlab{b}}.
\newblock URL \url{https://qwen.ai/blog?id=qwen3.6-27b}.

\bibitem[Rawles et~al.(2025)Rawles, Clinckemaillie, Chang, Waltz, Lau, Fair,
  Li, Bishop, Li, Campbell-Ajala, et~al.]{rawles2025androidworld}
Chris Rawles, Sarah Clinckemaillie, Yifan Chang, Jonathan Waltz, Gabrielle Lau,
  Marybeth Fair, Alice Li, William Bishop, Wei Li, Folawiyo Campbell-Ajala,
  et~al.
\newblock Androidworld: A dynamic benchmarking environment for autonomous
  agents.
\newblock In \emph{International Conference on Learning Representations},
  volume 2025, pages 406--441, 2025.

\bibitem[Shen et~al.(2026)Shen, Li, Zhou, and Hu]{shen2026mem2actbench}
Yiting Shen, Kun Li, Wei Zhou, and Songlin Hu.
\newblock Mem2actbench: A benchmark for evaluating long-term memory utilization
  in task-oriented autonomous agents.
\newblock In \emph{Proceedings of the 64th Annual Meeting of the Association
  for Computational Linguistics (Volume 1: Long Papers)}, pages 8173--8190,
  2026.

\bibitem[Shinn et~al.(2023)Shinn, Cassano, Gopinath, Narasimhan, and
  Yao]{shinn2023reflexion}
Noah Shinn, Federico Cassano, Ashwin Gopinath, Karthik Narasimhan, and Shunyu
  Yao.
\newblock Reflexion: Language agents with verbal reinforcement learning.
\newblock \emph{Advances in neural information processing systems},
  36:\penalty0 8634--8652, 2023.

\bibitem[Tavakoli et~al.(2026)Tavakoli, Salemi, Ye, Abdalla, Zamani, and
  Mitchell]{tavakoli2026beyond}
Mohammad Tavakoli, Alireza Salemi, Carrie Ye, Mohamed Abdalla, Hamed Zamani,
  and J.~Ross Mitchell.
\newblock Beyond a million tokens: Benchmarking and enhancing long-term memory
  in llms.
\newblock In \emph{International Conference on Learning Representations},
  volume 2026, pages 133343--133429, 2026.

\bibitem[{Venus Team} et~al.(2026){Venus Team}, Cai, Chen, Chen, Chen, Gao, Gu,
  Guo, Hu, Jiang, et~al.]{team2026ui}
{Venus Team}, Zhuohan Cai, Haoxing Chen, Jiaxuan Chen, Weizhi Chen, Changlong
  Gao, Zhangxuan Gu, Yuan Guo, Yusong Hu, Jianrong Jiang, et~al.
\newblock Ui-venus-2 technical report.
\newblock \emph{arXiv preprint arXiv:2609.00028}, 2026.

\bibitem[Wang et~al.(2021)Wang, Li, Zhao, and Yu]{wang2021naturalconv}
Xiaoyang Wang, Chen Li, Jianqiao Zhao, and Dong Yu.
\newblock Naturalconv: A chinese dialogue dataset towards multi-turn
  topic-driven conversation.
\newblock In \emph{Proceedings of the AAAI Conference on Artificial
  Intelligence}, volume~35, pages 14006--14014, 2021.

\bibitem[Wu et~al.(2024)Wu, Wang, Yu, Zhang, Chang, and Yu]{wu2024longmemeval}
Di~Wu, Hongwei Wang, Wenhao Yu, Yuwei Zhang, Kai-Wei Chang, and Dong Yu.
\newblock Longmemeval: Benchmarking chat assistants on long-term interactive
  memory.
\newblock \emph{arXiv preprint arXiv:2410.10813}, 2024.

\bibitem[Wu et~al.(2026)Wu, Hao, Wang, Wu, Xiao, Li, Zhou, Ju, Liu, Fan,
  et~al.]{wu2026mobilegym}
Dingbang Wu, Rui Hao, Haiyang Wang, Shuzhe Wu, Han Xiao, Zhenghong Li, Bojiang
  Zhou, Zheng Ju, Zichen Liu, Lue Fan, et~al.
\newblock Mobilegym: A verifiable and highly parallel simulation platform for
  mobile gui agent research.
\newblock \emph{arXiv preprint arXiv:2605.26114}, 2026.

\bibitem[Xiao et~al.(2026)Xiao, Yu, Wang, Zheng, Lin, Liu, Ding, Zhang, Wang,
  Feng, et~al.]{xiao2026alpsbench}
Jianfei Xiao, Xiang Yu, Chengbing Wang, Wuqiang Zheng, Xinyu Lin, Kaining Liu,
  Hongxun Ding, Yang Zhang, Wenjie Wang, Fuli Feng, et~al.
\newblock Alpsbench: An llm personalization benchmark for real-dialogue
  memorization and preference alignment.
\newblock \emph{arXiv preprint arXiv:2603.26680}, 2026.

\bibitem[Xu et~al.(2025)Xu, Liang, Mei, Gao, Tan, and Zhang]{xu2025amem}
Wujiang Xu, Zujie Liang, Kai Mei, Hang Gao, Juntao Tan, and Yongfeng Zhang.
\newblock {A-MEM}: Agentic memory for {LLM} agents.
\newblock In \emph{Advances in Neural Information Processing Systems},
  volume~38, 2025.
\newblock URL
  \url{https://proceedings.neurips.cc/paper_files/paper/2025/hash/19909c36f51abc4856b4560aff3d36d6-Abstract-Conference.html}.

\bibitem[Zhang et~al.(2026)Zhang, Tong, Fu, Zhao, Jiang, Jiang, and
  Yang]{zhang2026lunar}
Jiahao Zhang, Yongzhi Tong, Zelin Fu, Pengde Zhao, Yanmei Jiang, Feng Jiang,
  and Min Yang.
\newblock Lunar: Benchmarking personalized large language models on universal
  user behavior logs.
\newblock \emph{arXiv preprint arXiv:2608.05246}, 2026.

\bibitem[Zhang et~al.(2025)Zhang, Li, Long, Zhang, Lin, Yang, Xie, Yang, Liu,
  Lin, Huang, and Zhou]{qwen3embedding}
Yanzhao Zhang, Mingxin Li, Dingkun Long, Xin Zhang, Huan Lin, Baosong Yang,
  Pengjun Xie, An~Yang, Dayiheng Liu, Junyang Lin, Fei Huang, and Jingren Zhou.
\newblock Qwen3 embedding: Advancing text embedding and reranking through
  foundation models.
\newblock \emph{arXiv preprint arXiv:2506.05176}, 2025.

\bibitem[Zhao et~al.(2024)Zhao, Huang, Xu, Lin, Liu, and Huang]{zhao2024expel}
Andrew Zhao, Daniel Huang, Quentin Xu, Matthieu Lin, Yong-Jin Liu, and Gao
  Huang.
\newblock Expel: Llm agents are experiential learners.
\newblock In \emph{Proceedings of the AAAI Conference on Artificial
  Intelligence}, volume~38, pages 19632--19642, 2024.

\bibitem[Zhong et~al.(2024)Zhong, Guo, Gao, Ye, and Wang]{zhong2024memorybank}
Wanjun Zhong, Lianghong Guo, Qiqi Gao, He~Ye, and Yanlin Wang.
\newblock Memorybank: Enhancing large language models with long-term memory.
\newblock In \emph{Proceedings of the AAAI conference on artificial
  intelligence}, volume~38, pages 19724--19731, 2024.

\bibitem[Zhu et~al.(2026)Zhu, Wu, Wang, Qu, Liu, Cao, Chen, Sun, Zhu, Zhong,
  et~al.]{zhu2026mobilepa}
Yi~Zhu, Xiongwei Wu, Qiyi Wang, Tingyu Qu, Jiajun Liu, Sihan Cao, Long Chen,
  Weigao Sun, Feida Zhu, Yiran Zhong, et~al.
\newblock Mobilepa-bench: Benchmarking mobile planner agents on complex
  real-world tasks.
\newblock \emph{arXiv preprint arXiv:2608.23035}, 2026.

\end{thebibliography}
\bibliographystyle{plainnat}

\clearpage
\appendix

\section*{Appendix Overview}
Appendix~\ref{app:construction} describes how the ten worlds are planned, realized as application records, grounded into tasks, and validated.
Appendix~\ref{app:benchmark_details} reports dataset statistics, the two task labels, representative tasks, and a comparison of scale with prior benchmarks.
Appendix~\ref{app:experimental_setup} specifies memory formation, the eight memory backends, the evaluation settings and execution protocol, and task scoring.
Appendix~\ref{app:tools_prompts} lists the GUI and search tools and gives excerpts of the agent, construction, and evaluation prompts.
Appendix~\ref{app:limitations} discusses the limitations of the benchmark.
\section{Benchmark Construction and Validation}
\label{app:construction}

\subsection{Construction Implementation and Models}

Section~\ref{sec:r2forge} outlines the three stages of $R^2$-Forge.
During world planning, dependencies between events are represented as a directed acyclic graph, where an edge $e_i \rightarrow e_j$ indicates that event $e_j$ builds on state or context established by $e_i$, and sessions are scheduled onto the timeline $H=(h_1,\ldots,h_n)$ under these constraints.
During state realization, conversational records are generated with prompt designs informed by dialogue corpora~\citep{wang2021naturalconv,li2017dailydialog}, personal images are synthesized from session-conditioned scene descriptions, and application screenshots are rendered directly from MobileGym states.
Table~\ref{tab:construction_models} lists the models used at each construction and review stage, and Appendix~\ref{app:construction_prompts} provides prompt excerpts.

\begin{table}[!htbp]
\centering
\tablestyle
\setlength{\tabcolsep}{7pt}
\caption{Models used in benchmark construction and validation.}
\label{tab:construction_models}
\begin{tabular*}{\textwidth}{@{\extracolsep{\fill}}ll@{\hspace{1.8em}}ll@{}}
\toprule
\textbf{Stage / Role} & \textbf{Model} & \textbf{Stage / Role} & \textbf{Model} \\
\cmidrule(r){1-2}\cmidrule(l){3-4}
World planning and task authoring & GPT-5.6-sol & Image generation & GPT Image 2 \\
Record generation and grounding & Gemini 3.1 Pro Preview & Captioning and OCR & Gemini 3.6 Flash \\
Stage review & Claude Opus 5 & Image and visual-fact review & Gemini 3.6 Flash \\
Final task source review & Gemini 3.6 Flash & Semantic field validation & Qwen3.6-27B \\
\bottomrule
\end{tabular*}
\end{table}

\subsection{Task Validation Procedure}
\label{app:validation}

We validate each task for source grounding, evidence dependency, and executable outcome.

\paragraph{Source grounding.}
Each task is independently checked against the frozen realized records.
The review verifies that the target outcome, temporal relations, entity
references, and designated evidence are supported by the underlying world.
Reviewers also inspect the current environment for alternative support,
especially when a task is intended to depend on historical information that
is no longer available in $E_T$.

\paragraph{Evidence dependency.}
For each task $q_i$, let
\[
G_i=\{g_{i1},\ldots,g_{ik_i}\}
\]
denote its designated evidence.
For targeted counterfactual checks, we additionally define removal sets
\[
C_i=\{C_{ij}\}_{j=1}^{r_i},
\qquad
C_{ij}\subseteq G_i .
\]
We examine the task under three evidence settings:
\[
G_i,
\qquad
\emptyset,
\qquad
\{G_i\setminus C_{ij}\}_{j=1}^{r_i}.
\]
The full-evidence setting checks that the intended support is sufficient.
The empty-evidence setting checks whether the target remains recoverable
without that support.
Targeted removals test whether deleting task-critical evidence changes
answerability as annotated.

\paragraph{Executable outcome.}
Every task is evaluated with a terminal-state rubric that specifies the
required outcome, permitted changes, and applicable state-preservation
constraints.
For each rubric, we instantiate a terminal state that satisfies the task and
verify that it is accepted.
We also evaluate the unchanged initial state to ensure that the task is not
credited without realizing the requested outcome.
Together, these checks connect grounded evidence to a verifiable executable
result.

\section{Benchmark Composition and Task Design}
\label{app:benchmark_details}

\subsection{World and Task Composition}
\label{app:dataset_composition}
\label{app:task_dimensions}

\begin{table}[!htbp]
\centering
\tablestyle
\setlength{\tabcolsep}{8pt}
\caption{Dataset composition across the ten user worlds. Means weight the ten worlds equally.}
\label{tab:dataset_statistics}
\begin{tabular*}{\textwidth}{@{\extracolsep{\fill}}lr@{\hspace{2.5em}}lr@{}}
\toprule
\textbf{Statistic} & \textbf{Value} & \textbf{Statistic} & \textbf{Value} \\
\cmidrule(r){1-2}\cmidrule(l){3-4}
Worlds (Chinese / English) & 10 (5 / 5) & Records per world, mean & 8,746 \\
Tasks (Chinese / English) & 513 (254 / 259) & Records per world, range & 7,757--11,567 \\
Applications covered & 23 & History span per world (days) & 462--547 \\
Application records & 87,459 & Mean record share for $M_\tau$ (\%) & 56.5 \\
\bottomrule
\end{tabular*}
\end{table}

\begin{figure}[H]
\centering
\includegraphics[width=\textwidth]{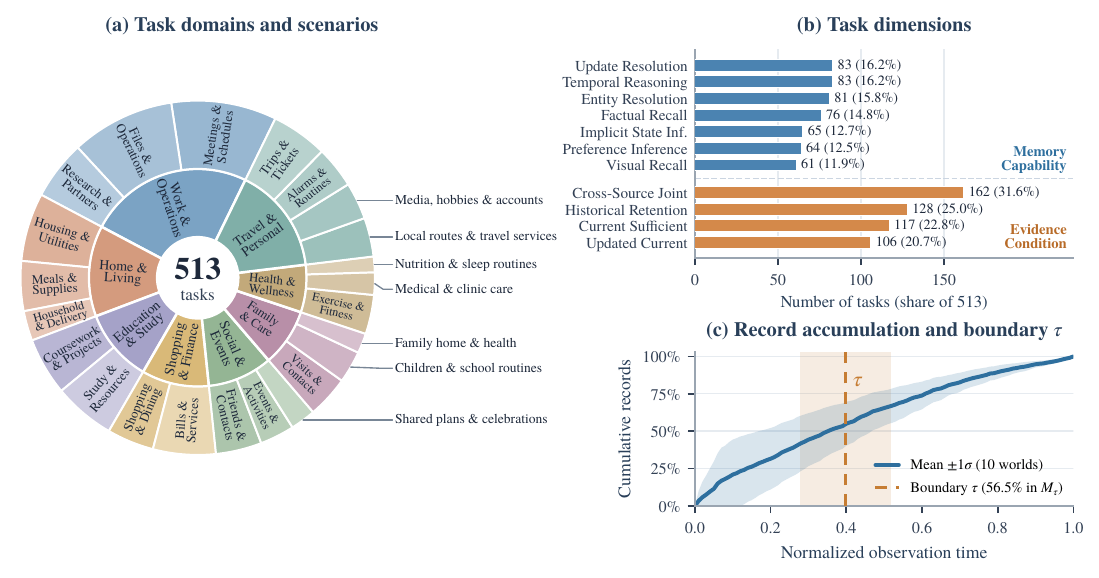}
\caption{\textbf{Benchmark composition and record accumulation across the ten worlds.}
(a) Distribution of the 513 tasks across life domains (inner ring) and scenarios (outer ring).
(b) Task counts and shares by Memory Capability and Evidence Condition.
(c) Mean cumulative record share and memory boundary $\tau$ across the ten worlds, with shaded bands showing $\pm 1\sigma$.}
\label{fig:record_history}
\end{figure}

Table~\ref{tab:dataset_statistics} and Figure~\ref{fig:record_history} summarize the benchmark composition across the ten worlds.
Figure~\ref{fig:record_history}a groups the 513 tasks by the life domains and scenarios of the events from which they are derived, and Figure~\ref{fig:record_history}b reports their distribution across the seven Memory Capabilities and four Evidence Conditions.
Capability counts range from 61 to 83 tasks (11.9--16.2\%), while Historical Retention and Cross-Source Joint together account for 290 tasks (56.5\%) whose required historical records are no longer fully available in the current environment.
Figure~\ref{fig:record_history}c shows the mean cumulative record distribution across the ten worlds after normalizing time and record counts within each world, where the memory boundary $\tau$ falls at 39.8\% of the observation window on average and includes 56.5\% of the records.

\subsection{Representative Task Cases}
\label{app:task_cases}
\label{app:capability_cases}
\label{app:evidence_cases}

Memory Capability specifies what operation is required over user-specific information, whereas Evidence Condition specifies where the required records reside at evaluation time.
Table~\ref{tab:representative_tasks} illustrates all seven Memory Capabilities and four Evidence Conditions with representative tasks from the English worlds.

\begin{table}[H]
\centering
\footnotesize
\renewcommand{\arraystretch}{1.12}
\setlength{\tabcolsep}{6pt}
\caption{Representative tasks across the seven Memory Capabilities and four Evidence Conditions.}
\label{tab:representative_tasks}
\begin{tabular}{@{}>{\raggedright\arraybackslash}p{0.19\textwidth}>{\raggedright\arraybackslash}p{0.43\textwidth}>{\raggedright\arraybackslash}p{0.33\textwidth}@{}}
\toprule
\textbf{Task Dimensions} & \textbf{User Request} & \textbf{Grounded Records and Resolution} \\
\midrule
\textbf{Factual Recall}\newline\textit{Historical Retention}
& ``Set my recurring Work alarm back to the time it had on February 3. Keep it enabled and preserve all its current settings other than the time.''
& The February 3 alarm time (\textbf{8:00 AM}) is no longer visible in $E_T$. The agent must recover this historical value from $M_\tau$ and restore the alarm in Clock. \\
\addlinespace[3pt]
\textbf{Update Resolution}\newline\textit{Updated Current}
& ``Add one line to the end of the notes of my next scooter routine inspection in the calendar (keep the existing text): the tire pressure and the long-term storage battery charge range that my scooter care archive's master index recommends now. Don't change anything else in my calendar.''
& In June 2026, the scooter care index raised the recommended tire pressure from 2.2 bar to \textbf{2.5 bar} and changed the storage charge range from 20\%--80\% to \textbf{40\%--60\%}. The agent must use the updated index in $E_T$ rather than the earlier version. \\
\addlinespace[3pt]
\textbf{Temporal Reasoning}\newline\textit{Cross-Source Joint}
& ``For my catch-up with Han Anran, what was the original tentative date, what was the finally confirmed date, and how many calendar days apart were they?''
& The tentative date (\textbf{October 11, 2025}) is recorded before $\tau$ and absent from $E_T$, whereas the confirmed date (\textbf{January 10, 2026}) is entered after $\tau$. Combining both sources yields \textbf{91 calendar days}. \\
\addlinespace[3pt]
\textbf{Entity Resolution}\newline\textit{Updated Current}
& ``Where does the person who asked me to come on Saturday instead because Sunday would be too busy currently work?''
& A WeChat message identifies \textbf{Kong Siyan}, whose contact entry was updated after $\tau$ in $E_T$ to \textbf{Qiantang Education Bureau}. \\
\addlinespace[3pt]
\textbf{Preference Inference}\newline\textit{Current Sufficient}
& ``Schedule a 45-minute Third Work Review on August 1, 2026, at the time I usually use for work review meetings.''
& Previous work-review meetings remaining in $E_T$ repeatedly start at \textbf{10:30 AM}. The agent must infer this recurring pattern and schedule the new calendar event. \\
\addlinespace[3pt]
\textbf{Implicit State Inference}\newline\textit{Current Sufficient}
& ``Across the September 2, 2025 Q2 correction and the May 15, 2026 billing period, what prerequisite do I appear to use before paying a property-fee notice?''
& Comparing the two billing episodes in $E_T$ derives the unstated user prerequisite \textbf{verify the statement before paying}. \\
\addlinespace[3pt]
\textbf{Visual Recall}\newline\textit{Current Sufficient}
& ``In the CloudPurse chat where you supplied the requested 15-digit order number for an electronic refund receipt, look at the receipt image sent immediately afterward. Create only one note titled `Refund Receipt Image' that gives the order number I sent, the color of the receipt paper, the icon shown below the receipt (its color and shape), and how many sheets of paper the image shows.''
& The chat context in $E_T$ identifies the receipt image, from which the agent recovers order \textbf{CP202505248839215}, \textbf{white} paper, a \textbf{green circular check-mark icon}, and \textbf{one} sheet. \\
\bottomrule
\end{tabular}
\end{table}

\subsection{Extended Comparison with Existing Benchmarks}
\label{app:dataset_comparison}

Table~\ref{tab:benchmark_scale} extends Table~\ref{tab:benchmark_comparison} with dataset size, task type, language, and token history length.
History lengths count the text and image descriptions in each history using \texttt{tiktoken} (\texttt{o200k\_base}).

\begin{table}[H]
\centering
\tablestyle
\setlength{\tabcolsep}{3pt}
\caption{\textbf{Benchmark scale and history length.} History lengths are measured in tokens.}
\label{tab:benchmark_scale}
\resizebox{\textwidth}{!}{\begin{tabular}{@{}lrrlclrrr@{}}
\toprule
\multirow{2}{*}[-0.45ex]{\textbf{Benchmark}} & \multirow{2}{*}[-0.45ex]{\textbf{Histories}} & \multirow{2}{*}[-0.45ex]{\textbf{Tasks}} & \multirow{2}{*}[-0.45ex]{\textbf{Task type}} & \multirow{2}{*}[-0.45ex]{\textbf{Language}} & \multirow{2}{*}[-0.45ex]{\textbf{History unit}} & \multicolumn{3}{c}{\textbf{History length}} \\
\cmidrule(lr){7-9}
 & & & & & & \textbf{Min.} & \textbf{Mean} & \textbf{Max.} \\
\midrule
LoCoMo-10 & 10 & 1,986 & QA & EN & Conversation & 13k & 21k & 25k \\
LongMemEval-S & 500 & 500 & QA & EN & Query context & 97k & 103k & 106k \\
BEAM-100K & 20 & 400 & QA & EN & Conversation & 94k & 116k & 123k \\
BEAM-500K & 35 & 700 & QA & EN & Conversation & 413k & 506k & 718k \\
BEAM-1M & 35 & 700 & QA & EN & Conversation & 807k & 960k & 1.0M \\
BEAM-10M & 10 & 200 & QA & EN & Conversation & 9.6M & 9.7M & 9.8M \\
\addlinespace[4pt]
Mem2ActBench & 2,029 & 400 & Tool use & EN & Session & 1k & 3k & 16k \\
MemoryArena & --- & 701 & Interactive & EN & Online rollout & --- & --- & --- \\
WorldMemArena & 461 & 461 & QA & EN & Sample history & 1k & 12k & 90k \\
\addlinespace[4pt]
MobileMem (Text) & 2 & 1,319 & QA & ZH & User history & 186k & 227k & 269k \\
MobileMem (Omni) & 16 & 7,415 & QA & EN/ZH & User history & --- & 1.7M\textsuperscript{*} & --- \\
MemGUI-Bench & --- & 128 & GUI & EN & Online rollout & --- & --- & --- \\
KnowU-Bench & 4 & 192 & GUI & ZH & Profile log & 6k & 9k & 12k \\
\midrule
\rowcolor{PaperPanel}
\textbf{$\bm{R^2}$-MobileMemory} & 10 & 513 & GUI & EN/ZH & World history & 497k & 628k & 991k \\
\bottomrule
\end{tabular}}
\par\smallskip
\begin{minipage}{\textwidth}
\footnotesize\raggedright
EN: English; ZH: Chinese. A dash indicates an unavailable value or no fixed history.\\
\textsuperscript{*}Author-reported mean from MobileMem Table~3~\citep{deng2026mobilemem}; all other non-empty lengths use \texttt{o200k\_base}.
\end{minipage}
\end{table}

\section{Experimental Protocol}
\label{app:experimental_setup}

\subsection{Memory Formation}
Backends ingest one ordered stream of application records, using a common textual representation of record content and the visual facts that passed review, with markers indicating when a record is edited.
This shared representation enables comparison across backends with different modality support, similar to the caption projection in MobileMem-Omni~\citep{deng2026mobilemem}.
For each world, we save $M_\tau=\mathrm{Ingest}(R_{\leq\tau})$ at the memory boundary and continue ingestion through evaluation time to obtain $M_T=\mathrm{Ingest}(R_{\leq T})$.
Backends that require an LLM or an embedding model use Qwen3.6-27B~\citep{qwen36_27b} and Qwen3-Embedding-8B~\citep{qwen3embedding}.

\subsection{Memory Backends}
\label{app:memory_backends}

All eight backends ingest this record stream and expose retrieval through \texttt{memory\_search}. They differ in how they organize incoming records and retrieve evidence for the agent.

\begin{itemize}
\setlength{\itemsep}{6pt}
\item \textbf{NaiveRAG} is a dense-retrieval baseline following retrieval-augmented generation~\citep{lewis2020rag}. It stores the original record text in an embedding index and returns the memory units most relevant to each query.

\item \textbf{HippoRAG2}~\citep{gutierrez2025rag} builds a graph of extracted phrases and source passages. Dense retrieval and LLM-based relevance filtering identify query-relevant evidence, which initializes Personalized PageRank over the graph. The resulting ranking retrieves passages through associative connections.

\item \textbf{A-MEM}~\citep{xu2025amem} organizes records as interconnected notes with contextual descriptions, keywords, and tags. During ingestion, it links new notes to related memories and updates their contextual attributes, allowing the memory structure to evolve with the record stream.

\item \textbf{MemOS}~\citep{li2025memos} manages memory through MemCubes that package content with provenance and lifecycle metadata. We use its \texttt{TreeTextMemory} component to organize textual records and retrieve relevant memory content.

\item \textbf{EverMemOS}~\citep{hu2026evermemos} extracts episodic MemCells containing facts and temporally scoped information, then consolidates related cells into thematic MemScenes. We use hybrid retrieval over this memory.

\item \textbf{Mem0 and Mem0$^{g}$}~\citep{chhikara2025mem0} maintain fact-based memory. Mem0 extracts key facts from incoming records and reconciles them with existing memory through addition, update, and deletion. Mem0$^{g}$ adds entities and relations to this fact store and combines graph-based context with vector retrieval. We evaluate each configuration as a separate backend.

\item \textbf{LangMem}~\citep{langchain2025langmem} uses an LLM-based memory manager to extract, consolidate, and update persistent information. Its memory-store manager maintains record-derived memories in an embedding-indexed store and retrieves relevant entries for each query.
\end{itemize}

\subsection{Agent Execution Protocol}
\label{app:execution_protocol}

Each task runs independently in MobileGym starting from $E_T$, where the agent receives the user request, the current screenshot, and the tools available under its assigned setting for up to 50 interaction steps.
One step corresponds to one model response, in which the agent may issue multiple retrieval calls or a single GUI action, but cannot combine both in the same step.
Memory retrieval returns at most ten entries per \texttt{memory\_search} call, and memory is not updated during execution.
Detailed tool schemas and agent instructions are provided in Appendix~\ref{app:tools_prompts}.

\subsection{Task Evaluation}
\label{app:evaluation_protocol}

The evaluator verifies the terminal MobileGym state rather than an agent's declaration of completion.
Structured state properties are checked deterministically, while criteria requiring semantic equivalence are judged against the expected values and observed terminal-state content, combined with applicable authorization and side-effect constraints.
A run terminates when the agent calls \texttt{gui\_complete} or \texttt{gui\_abort}, or reaches the 50-step limit.
Prompt excerpts for semantic outcome evaluation are provided in Appendix~\ref{app:construction_prompts}.

\section{Tool Interfaces and Prompt Specifications}
\label{app:tools_prompts}

\subsection{Tool Interfaces}
\label{app:tool_interfaces}

Tables~\ref{tab:gui_tools} and~\ref{tab:retrieval_tools} list the GUI control tools and retrieval tools.
Application search combines text matching with application-specific filters, and the agent can open any result with \texttt{phone\_read\_result} to inspect full text, surrounding conversational context, or attached images.

\begin{table*}[!htbp]
\centering
\footnotesize
\renewcommand{\arraystretch}{1.12}
\setlength{\tabcolsep}{3.0pt}
\begin{minipage}[t]{0.39\textwidth}
\centering
\caption{\textbf{GUI control tools.}}
\label{tab:gui_tools}
\begin{tabular*}{\linewidth}{@{\extracolsep{\fill}}ll@{}}
\toprule
\textbf{Tool} & \textbf{Action} \\
\midrule
\texttt{gui\_click} & Tap a location \\
\texttt{gui\_double\_tap} & Double-tap a location \\
\texttt{gui\_long\_press} & Press and hold \\
\texttt{gui\_type} & Enter text; optionally clear \\
\texttt{gui\_swipe} & Swipe between locations \\
\texttt{gui\_drag} & Drag between locations \\
\texttt{gui\_back} & Go back \\
\texttt{gui\_home} & Return to home screen \\
\texttt{gui\_recent} & Show recent apps \\
\texttt{gui\_enter} & Press Enter \\
\texttt{gui\_wait} & Wait for 0--10 seconds \\
\texttt{gui\_awake} & Open an app by name \\
\texttt{gui\_complete} & Mark completion \\
\texttt{gui\_abort} & Stop and give reason \\
\bottomrule
\end{tabular*}
\end{minipage}\hfill
\begin{minipage}[t]{0.58\textwidth}
\centering
\caption{\textbf{Retrieval tools.}}
\label{tab:retrieval_tools}
\resizebox{\linewidth}{!}{\begin{tabular}{@{}ll@{}}
\toprule
\textbf{Tool} & \textbf{Inputs} \\
\midrule
\texttt{memory\_search} & Query; up to 10 memory entries \\
\texttt{wechat\_search\_messages} & Query, chat, sender, type, direction, time \\
\texttt{sms\_search\_messages} & Query, counterparty, phone, direction, time \\
\texttt{calendar\_search\_events} & Query, location, start time \\
\texttt{tencent\_meeting\_search\_meetings} & Query, kind, host, status, start time \\
\texttt{contacts\_search} & Query, scope, call direction, call time \\
\texttt{notes\_search\_notes} & Query, folder, update time \\
\texttt{file\_manager\_search\_files} & Query, path, media type, modified time \\
\texttt{gallery\_search\_media} & Query, album, media type, saved time \\
\texttt{railway12306\_search\_orders} & Train, stations, passenger, status, date \\
\texttt{phone\_read\_result} & Result ref; offset, limit, context, crop \\
\bottomrule
\end{tabular}}
\end{minipage}
\end{table*}
\FloatBarrier

\subsection{Agent Prompt}
\label{app:agent_prompt}

Table~\ref{tab:agent_prompt} provides excerpts of the agent instructions
implementing the execution protocol in Appendix~\ref{app:execution_protocol},
grouped by setting and followed by a summary of task-completion
requirements.

\begin{table}[!htbp]
\centering
\tablestyle
\renewcommand{\arraystretch}{1.06}
\setlength{\tabcolsep}{8pt}
\caption{\textbf{Agent instructions.} Prompt excerpts grouped by setting, followed by a summary of completion requirements.}
\label{tab:agent_prompt}
\begin{tabular}{>{\raggedright\arraybackslash}p{\dimexpr\linewidth-16pt\relax}}
\toprule
\rowcolor{TableGroup}
\rule{0pt}{2.4ex}\textcolor{PaperInk}{\textbf{Role and interaction}}\hfill\textit{Search-enabled settings} \\
You are a mobile-phone agent. Retrieve information and operate the current phone only through the registered function tools. You may call multiple read-only retrieval tools in one turn. GUI tools change the screen, so call at most one GUI tool per turn and do not combine it with retrieval tools in the same turn. \\
\addlinespace[2pt]
The conversation retains your previous tool calls and their results; only the current screen screenshot is kept. Search returns \texttt{result\_ref} values. Use \texttt{phone\_read\_result} to read full text, nearby messages, or actual images; crop images to inspect details when needed. \\
\addlinespace[4pt]
\rowcolor{TableGroup}
\rule{0pt}{2.4ex}\textcolor{PaperInk}{\textbf{Memory sources}}\hfill\textit{Memory settings} \\
\texttt{memory\_search} returns long-term memories formed from past experiences. Phone records may have been updated or overwritten since those memories were formed, and the phone displays only its current state. Earlier historical information may therefore exist only in memory, while recent changes may exist only on the phone. Both are available read-only evidence sources, and neither is guaranteed to be sufficient on its own. Choose sources according to the question. \\
\addlinespace[2pt]
\texttt{memory\_search} returns up to 10 memories per call. The retrieval count is fixed by the evaluation; provide only query and choose its wording for the question. \\
\addlinespace[4pt]
\rowcolor{TableGroup}
\rule{0pt}{2.4ex}\textcolor{PaperInk}{\textbf{Role and access}}\hfill\textit{GUI Only} \\
You are a mobile-phone agent. Operate the current phone through the registered GUI function tools. Call at most one GUI tool per turn. Read information on the screen; you may also use search interfaces inside apps. \\
\addlinespace[4pt]
\rowcolor{TableGroup}
\rule{0pt}{2.4ex}\textcolor{PaperInk}{\textbf{Completion and output}}\hfill\textit{Summary} \\
Verify the information needed for the user request, then use the GUI to realize the required terminal state. Follow the task-specific completion instruction and modify only authorized targets. Coordinates are normalized to 0--1000 in $(x,y)$ order. Preserve exact names, source quotations, numbers, and required field values. \\
\bottomrule
\end{tabular}
\end{table}
\FloatBarrier

\subsection{Construction and Evaluation Prompt Excerpts}
\label{app:construction_prompts}

Table~\ref{tab:construction_prompts} presents edited English excerpts of the instructions used for fact extraction, task authoring, independent source review, and semantic outcome evaluation.

\begin{table}[!htbp]
\centering
\tablestyle
\renewcommand{\arraystretch}{1.01}
\setlength{\tabcolsep}{8pt}
\caption{\textbf{Construction and evaluation instructions.} Edited prompt excerpts with the inputs and outputs of each stage.}
\label{tab:construction_prompts}
\begin{tabular}{>{\raggedright\arraybackslash}p{\dimexpr\linewidth-16pt\relax}}
\toprule
\rowcolor{TableGroup}
\rule{0pt}{2.4ex}\textcolor{PaperInk}{\textbf{Fact extraction}} \\
Extract one to six atomic facts from the reviewed, user-visible records in the current session. Each fact must express one independently updateable property, identify its subject and value, and cite supporting records. Cover the supplied Fact Intents only when realized in the records. Preserve uncertainty and pending status; use the latest supported state within the session. Leave cross-session version linking to the subsequent stage. \\
\addlinespace[2pt]
\textit{Input:} Session records, subject catalog, Fact Intents, and review feedback.\\
\textit{Output:} structured facts with values, dimensions, and source references. \\
\addlinespace[2pt]
\rowcolor{TableGroup}
\rule{0pt}{2.3ex}\textcolor{PaperInk}{\textbf{Task authoring}} \\
Construct one executable mobile task from the supplied frozen records. Focus on one coherent user goal. Preserve record identities, timestamps, and the execution contract. Ground every target value in source evidence; do not reveal it in the user request. Specify the required terminal state, authorized changes, and state-preservation constraints. Assign the memory capability and evidence condition from the actual dependency on historical and current sources. \\
\addlinespace[1pt]
\textit{Input:} task blueprint and source packet.\\
\textit{Output:} user request, target contract, evidence references, and field-level source proofs. \\
\addlinespace[2pt]
\rowcolor{TableGroup}
\rule{0pt}{2.3ex}\textcolor{PaperInk}{\textbf{Independent source review}} \\
Check every target value against the frozen records. Verify temporal calculations, entity uniqueness, evidence sufficiency, and the permitted state changes. Search current records for alternative sources for information labeled as historical. Check that the assigned capability is required and that removing the designated evidence changes answerability as specified. Return accept, revise, or reject with concrete findings. \\
\addlinespace[1pt]
\textit{Input:} candidate task and independent access to its source packet.\\
\textit{Output:} source findings, evidence-condition assessment, and revision issues. \\
\addlinespace[2pt]
\rowcolor{TableGroup}
\rule{0pt}{2.3ex}\textcolor{PaperInk}{\textbf{Semantic outcome evaluation}} \\
Evaluate the observed result against the requested target and supplied criterion. Accept semantically equivalent expressions when they preserve the required entity, time, quantity, and meaning. Reject unsupported or conflicting content. Return a structured decision and its reason. \\
\addlinespace[1pt]
\textit{Input:} criterion, expected value, and observed terminal-state content.\\
\textit{Output:} criterion-level judgment. Deterministic checks and applicable authorization and side-effect checks are combined with these judgments to determine task success. \\
\bottomrule
\end{tabular}
\end{table}
\FloatBarrier

\section{Limitations}
\label{app:limitations}

\textbf{$\bm{R^2}$-MobileMemory} is built on ten simulated user worlds within MobileGym, so its application records and task distributions come from controlled synthesis and do not directly establish generalization to real users or physical devices.
In addition, to compare memory backends under a shared input protocol, memory formation uses a textual projection of application records and visual facts rather than native multimodal memory representations, although agents still inspect screenshots and application images during task execution.

\end{document}